%% file: 0.acl_latex.tex
\documentclass[11pt]{article}

\usepackage{times}
\usepackage{latexsym}

\usepackage{longtable}
\usepackage{subcaption}
\usepackage{float} 
\usepackage[final]{acl}
\usepackage{microtype}
\usepackage{inconsolata}
\usepackage{graphicx}
\usepackage{amsmath}
\usepackage{amssymb}
\usepackage{booktabs}
\usepackage{CJKutf8}

\title{Double-Edged Sword or Sharp Tool? Designing and Evaluating Triadic LLM-Teacher Collaboration for K-12 Writing at Scale}

\author{
 \textbf{Canran WANG\textsuperscript{1}$^{\#}$},
 \textbf{Yuwen Yang\textsuperscript{1}$^{\#}$},
 \textbf{Zhen Wang\textsuperscript{2}},
 \textbf{Ming MA\textsuperscript{3}},
\\
 \textbf{Chentai Wang\textsuperscript{3}},
 \textbf{Ding Yu\textsuperscript{3}},
 \textbf{Keman Huang\textsuperscript{1}\thanks{
The first two authors contributed equally to this work. Keman Huang is the corresponding author (keman@ruc.edu.cn).}},
 \textbf{Xiaoyong Du\textsuperscript{1}}
\\
\\
 \textsuperscript{1}Renmin University of China, Beijing, China, \\
 \textsuperscript{2}Central University of Finance and Economics, Beijing, China, 
\\
 \textsuperscript{3}Beijing HQ Intelligent Technology, Ltd., Beijing, China
}

\begin{document}
\maketitle

\begin{abstract}

The double-edged sword of integrating Large Language Models (LLMs) requires an effective triadic collaboration mechanism among LLMs, teachers and students, especially for K-12 education. By developing a triadic collaboration system to support K-12 writing learning, a multidimensional evaluation framework grounded in Systemic Functional Linguistics and the suggestion trajectory tracing pipeline, this paper contributes a large-scale empirical dataset involving $57,954$ essays from $10,195$ students across $120$ schools over two years. Our findings confirm the efficacy of this system in improving writing quality through a strategic labor division: the LLM serves as a generative engine to mitigate teacher burnout, and the teacher acts as a pedagogical gatekeeper and bridge to guarantee feedback quality. While both LLM and teacher are critical for skill improvement, we uncover a ceiling effect where excessive linguistic expansion yields diminishing marginal utility. These suggest a dynamically adaptive LLM-teacher collaboration as student proficiency increases.

\end{abstract}

\input{1.introduction}
\input{2.related-work}

\input{3.method}
\input{4.experiments-and-results}

\input{5.conclusion}
\input{6.limitation}

\input{7.ethical}

\section*{Acknowledgments}

The work was supported by the National Natural Science Foundation of China (62441230, 62172425), the Scientific Research Innovation Capability Support Project for Young Faculty (Grant No. SRICSPYF-ZY2025001), and the Fundamental Research Funds for the Central Universities and the Research Funds of Renmin University of China (22XNKJ04). 

\bibliography{custom}

\appendix
\input{appendix_annotation}

\end{document}

%% file: 1.introduction.tex
\section{Introduction}


The Large Language Models (LLMs) have fundamentally transformed educational practices\cite{florencio2025vibe}, supporting the entire learning process from curriculum design \cite{shojaei2025ai} to adaptive tutoring \cite{wang2024large}, particularly within higher education \cite{alghamdi2025leveraging}. However, a significant research gap persists in K-12 pedagogy: while LLMs lower the entry barrier for novice writers and enhance mnemonic retention through personalized interaction \cite{park2024empowering}, current studies predominantly target adult populations in controlled laboratory settings \cite{yuan2022wordcraft,jakesch2023co}, often overlooking the developmental nuances and socio-cognitive constraints of the K-12 classroom.

Crucially, LLM adoption presents a ``double-edged sword'' for developing learners: while these models can encourage divergent thinking \cite{wang2025large}, they risk homogenizing collective output \cite{doshi2024generative} and diminishing student ownership \cite{dhillon2024shaping,kadoma2024role}. Without human mediation, a reliance on algorithmic norms \cite{sundaresan2025algorithmic} threatens to stifle unique cultural voices and cognitive agency \cite{agarwal2025ai}. This suggests that LLMs alone are insufficient as partners for K-12 learners. Consequently, determining how LLMs and teachers can collaborate effectively to foster learning has emerged as a critical challenge for the application of NLP in pedagogy.

To bridge this gap, using K-12 writing as the research context, we design and implement a hybrid system featuring a \textbf{triadic collaboration} among students, LLMs, and teachers. This system synergizes the scalable generative power of LLMs with the pedagogical expertise of human educators, guiding students to transform their initial drafts into high-quality revisions. To evaluate this intervention, we introduce \textbf{a multidimensional evaluation framework} grounded in Systemic Functional Linguistics (SFL) \cite{halliday2013halliday}, capturing learning outcomes across the \textit{Ideational, Textual, and Interpersonal} metafunctions. Furthermore, we develop an analytical pipeline to \textbf{trace the trajectory} of how students incorporate specific suggestions into their final revisions. By deploying this system in real-world K-12 writing environments, this study asks:

\textbf{RQ1:} To what extent does the triadic collaboration system enhance student writing across the three SFL metafunctions?

\textbf{RQ2:} How is the labor divided between LLMs and teachers during feedback generation, and how are these distinct sources adopted by students?

\textbf{RQ3:} What are the underlying mechanisms through which hybrid LLM-teacher collaboration impacts writing performance?

To address these questions, we contribute \textbf{the first large-scale empirical study} centered on the K-12 iterative writing revision process for real-world K-12 writing environments. Our dataset spans over two years and comprises $N=57,954$ essays from $10,195$ students across $120$ schools. To the best of our knowledge, this is the most extensive empirical dataset in the domain of LLM-assisted K-12 education.

Furthermore, our findings provide systematic evidence for \textbf{the efficacy of triadic collaboration}, yielding a significant $5\%+$ improvement in rewriting quality. Beyond holistic gains cross all dimensions, our data reveals a pronounced growth in the interpersonal function, characterized by a distinct pro-social shift. Moreover, there exists \textbf{a strategic functional division of labor}: the LLM acts as a scalable generative engine to mitigate instructor burnout, and the teacher serves as a pedagogical gatekeeper and bridge to ensure feedback quality. Finally, while \textbf{both the LLM and teacher contribute significantly to performance, a critical ``ceiling effect''} emerges: although linguistic expansion initially proxies performance improvement, its marginal utility for the final grade decreases as proficiency rises, with certain metrics eventually turning negative. 
Overall, these findings highlight that effective LLM integration in K-12 requires a dynamic triadic design to ensure technological power is successfully converted into meaningful learning outcomes.

%% file: 2.related-work.tex
\section{Related Work}





\paragraph{LLMs in Education}
LLMs have been deeply embedded in educational ecosystems \cite{florencio2025vibe}, facilitating tasks ranging from mathematical reasoning \cite{wu2023mathchat} to the entire teaching-and-learning lifecycle\cite{chu2025LLM}. For educators, domain-specific agents serve as reflective planning tools that simulate classroom dynamics \cite{xu2025classroom}; for learners, they provide personalized feedback that bolsters motivation \cite{park2024empowering}. Despite these affordances, the proliferation of LLMs introduces critical challenges beyond academic integrity \cite{tang2024science}. Recent literature warns of cognitive deskilling, where outsourcing synthesis to algorithms leads to surface-level learning \cite{yang2026deskilling}. Moreover, over-reliance on LLMs may overdraw problem-solving capacities \cite{sundaresan2025algorithmic}, necessitating rigorous evaluation of AI's long-term pedagogical impact.


Critically, these conflicting outcomes remain unresolved due to a methodological reliance on adult cohorts in controlled settings (such as \cite{agarwal2025ai,jakesch2023co}, etc.). This demographic bias limits the generalizability of findings to K-12 contexts, necessitating empirical evidence that specifically targets the multidimensional developmental trajectories of developing writers.

\paragraph{Efficacy of LLM-Assisted Writing}
Current literature primarily frames LLM-assisted writing as a dyadic collaboration between a single user and an agent. While this partnership lowers entry barriers\cite{wang2024large} and boosts efficiency \cite{yuan2022wordcraft, agarwal2025ai, dhillon2024shaping}, it often results in a ``double-edged sword'' effect: quality gains often come at the expense of distinctiveness and psychological ownership \cite{dhillon2024shaping}. Critically, unmediated dyadic interaction risks linguistic homogenization \cite{guo2024curious, padmakumar2023does} and convergence toward Western-centric norms, potentially compelling minority users to abandon unique expressive styles \cite{jakesch2023co, agarwal2025ai, kadoma2024role}. Furthermore, these interventions may exacerbate inequalities, with outcomes varying significantly based on gender and cognitive capabilities \cite{lu2025we}.

Notably, existing research prioritizes dyadic ``Writer-LLM'' pairs in controlled settings, failing to capture the socially mentored nature of K-12 writing: efficacy is moderated by a \textbf{triadic collaboration} where teachers act as pedagogical mediators, filtering and contextualizing LLM outputs to align with learning objectives. Consequently, this study shifts the lens to this ``LLM-Teacher-Student'' collaboration to evaluate its impact on multidimensional linguistic development.

%% file: 3.method.tex
\section{Methodology}


\input{3.1.writing}

\input{3.2.metric}

\input{3.3.adoption}

%% file: 3.1.writing.tex
\begin{figure*}[t] 
    \centering
    \includegraphics[width=0.96\textwidth]{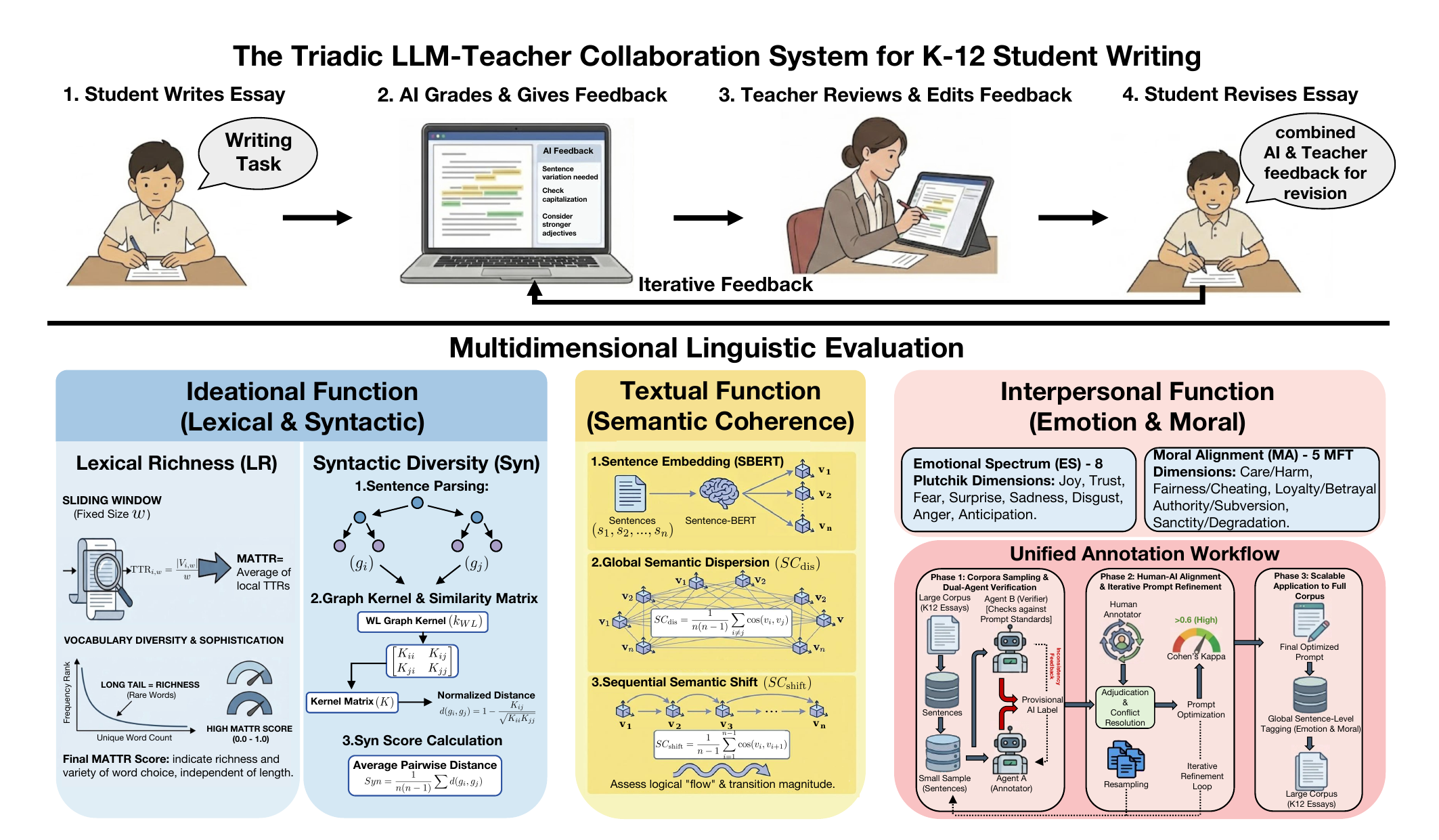}
    \caption{
        \textbf{The LLM-Teacher Collaboration System for K-12 Student Writing.} 
        The upper panel illustrates the iterative workflow where teacher expertise refines LLM-generated suggestions ($S_{initial}$) into actionable feedback ($S_{final}$) for student revisions. 
        The lower panel displays the evaluation metrics based on the Halliday's Systemic Functional Linguistics (SFL), covering Ideational ($Lex, Syn$), Textual ($SC_{dis},SC_{shift}$), and Interpersonal ($ES, MA$) linguistic dimensions.
        }
     \vspace{-1em}
    \label{fig:main_framework}
\end{figure*}

\subsection{The LLM-Teacher Collaboration System for K-12 Student Writing}

As illustrated in Figure~\ref{fig:main_framework}, we developed a hybrid system featuring a triadic collaboration between students, LLMs, and teachers. This architecture synergizes the scalable analytical power of LLMs with the pedagogical expertise of human educators to augment K-12 student writing skills. It transforms initial student drafts into enhanced compositions through a four-stage iterative process:

\textbf{Stage 1: Writing Task and Draft Generation.} The process initiates with students engaging in a standardized writing task tailored to K-12 curriculum standards. Students compose an initial draft, denoted as $D_{pre}$, which reflects their baseline linguistic and thematic proficiency. 

\textbf{Stage 2: LLM-Generated Feedback.} Upon submission, the initial draft ($D_{pre}$) is processed by the LLM, acting as an expert writing assistant. It rates the baseline writing quality ($\text{grade}_{\text{LLM}}^{pre}$) and generats a set of preliminary, constructive suggestions ($\mathcal{S}_{initial}$) across multiple dimensions. 

\textbf{Stage 3: Teacher Review and Refinement.} Rather than passing the LLM output directly to the student, the system incorporates a teacher who accesses the raw LLM suggestions ($\mathcal{S}_{initial}$) alongside the student's draft ($D_{pre}$). The teacher screens and refines the suggestions, including validating accurate insights, adjusting the tone, removing inappropriate or redundant outputs, adding missing pedagogical scaffolding to finalize the suggestions ($\mathcal{S}_{final}$) and providing the rating ($\text{grade}_{\text{teacher}}^{pre}$). 

\textbf{Stage 4: Feedback Adoption and Student Revision.} Upon receiving the teacher-mediated feedback ($\mathcal{S}_{\text{final}}$), students review and revise their original draft ($D_{\text{pre}}$) to produce a finalized version ($D_{\text{post}}$). To foster continuous improvement, this new draft cycles back for additional assessment, generating both the LLM evaluation ($\text{grade}_{\text{LLM}}^{\text{post}}$) and the teacher assessment ($\text{grade}_{\text{teacher}}^{\text{post}}$).

%% file: 3.2.metric.tex
\subsection{Multidimensional Linguistic Evaluation}

To provide a holistic assessment of student writing development, we ground our evaluation framework in Halliday's Systemic Functional Linguistics (SFL) \cite{halliday2013halliday}, which treats language as a social semiotic system where meaning-making occurs across three simultaneous functions (Ideational, Textual, and Interpersonal).


\subsubsection{Ideational Function}
The ideational function, measured by Lexical Richness ($Lex$) and Syntactic Diversity ($Syn$), quantifies the student's ability to deploy diverse vocabulary and complex grammatical structures.

\textbf{Lexical Richness ($Lex$)}: evaluates a student's expressive capacity and vocabulary breadth. For a given essay, we utilize the Moving-Average Type–Token Ratio (MATTR) \cite{agarwal2025ai} to calculate the Type–Token Ratio within a sliding window of fixed size $w$, which moves incrementally through the whole essay.

 \vspace{-1.5em}
\begin{equation} \text{MATTR} = \frac{1}{N - w + 1} \sum_{i=1}^{N - w + 1} {\frac{|V_{i, w}|}{w}} \end{equation}
 \vspace{-1em}

\noindent where $|V_{i, w}|$ is the number of unique tokens in the window starting at position $i$, $w$ is the window size and $N$ is the total number of tokens. A higher lexical richness score indicates higher richness and variety of word choice, typically associated with more advanced writing proficiency.


\textbf{Syntactic Diversity ($Syn$)}: We utilize a graph-based kernel approach \cite{guo2024curious} to capture the structural variation within each essay. Specifically, each sentence is parsed into a dependency graph $g$, where nodes represent words labeled with their part-of-speech (POS) tags and edges represent dependency relations \cite{che2021n}. For a set of $n$ sentence graphs, we calculate the Weisfeiler-Lehman (WL) graph kernel \cite{shervashidze2011weisfeiler} matrix $\mathbf{K}$ to derive the average pairwise distance between all sentences:

 \vspace{-1.5em}

\begin{equation}
{Syn} = \frac{1}{n(n-1)} \sum_{i \neq j} {(1 - \frac{k_{ij}}{\sqrt{k_{ii}k_{jj}}})}
\end{equation}
 \vspace{-1em}



\noindent where $k_{ij}$ is the structural similarity between the dependency graph for sentence $s_{i}$ and $s_{j}$ calculated by the WL kernel. A higher $Syn$ score indicates a broader range of grammatical structures and reduced syntactic monotonicity, typically representing a higher writing proficiency.

\subsubsection{Textual Function}

The textual function is operationalized through Semantic Dispersion ($SC_{dis}$) and Semantic Shift ($SC_{shift}$) to capture semantic coherence, which reflects how individual linguistic choices are woven into a logically and conceptually consistent student essay. In particular, we represent each sentence $s_i$ as a high-dimensional vector $\mathbf{v}_i$ using a pre-trained Sentence-BERT (SBERT) model \cite{reimers2019sentence} and then quantify semantic coherence through two distinct measures:

\textbf{Semantic Dispersion ($SC_{dis}$)}. We measure the overall breadth of ideas within an essay by calculating the average pairwise cosine distance between all sentence embeddings:

 \vspace{-0.5em}
\begin{equation}
\text{SC}_{dis} = \frac{1}{n(n-1)} \sum_{i \neq j} \cos(\mathbf{v}_i, \mathbf{v}_j)
\end{equation}
 \vspace{-0.5em}

\noindent where $n$ is the number of sentences and a higher $SC_{dis}$ indicates a broader semantic coverage and a more diverse range of topics within the essay.

\textbf{Semantic Shift ($SC_{shift}$)}. To assess the logical flow of the essay, we calculate the average distance between contiguous sentences:

\begin{equation}
\text{SC}_{shift} = \frac{1}{n-1} \sum_{i=1}^{n-1} \cos(\mathbf{v}_i, \mathbf{v}_{i+1})
\end{equation}

This sequential metric captures the magnitude of topical transitions. In the context of student writing, a very low $SC_{shift}$ score may indicate repetitive or redundant transitions. 


\subsubsection{Interpersonal Function}
The interpersonal function is measured using the Emotional Spectrum ($ES$) and Moral Alignment ($MA$) to evaluate students’ capacity to engage readers through pathos and ethical reasoning.

\textbf{Emotional Spectrum ($ES$)}. We ground our analysis in Plutchik’s wheel of emotions \cite{plaza2024emotion}, which identifies eight primary emotional dimensions in two sets: approach-oriented emotions including \textit{Anticipation, Joy, Trust, Fear} and avoidance-oriented emotions including \textit{Surprise, Sadness, Disgust, Anger}. To facilitate large-scale annotation, we developed a two-stage multi-agent pipeline utilizing the social-scientific labeling capabilities of LLMs. The process begins with iterative multi-agent discussions and human verification to optimize labeling prompts; our pilot validation yielded a Cohen's Kappa of 0.61, indicating strong alignment between the system and human experts. This optimized prompt was then deployed for large-scale automated annotation, where the Fleiss's Kappa for the final annotations reaches 0.981, confirming the high consistency for our labels. We provide the complete technical architecture, the finalized prompts, and the details of the consistency verification process in Appendix A.

Through this process, the emotional state of an essay is represented as an 8-dimensional vector: $\mathbf{e} = [e_1, \dots, e_8]$ where $e_i$ denotes the proportion of sentences categorized within a specific emotional dimension relative to the total sentence count $n$ of the essay. To quantify the balance of these emotions, we calculate the Emotional Spectrum ($ES$) using Shannon entropy:

 \vspace{-1em}

\begin{equation}
ES = -\sum_{i=1}^{8} \frac{e_i}{n} \log \frac{e_i}{n}
\end{equation}

A higher $ES$ score indicates a wider range of emotional registers, whereas a lower score suggests an emotionally flat or monotonic essay. 

\textbf{Moral Alignment ($MA$)}. Our analysis is based on the Moral Foundations Theory (MFT) \cite{abdulhai2024moral}, which identifies five positive ethics (\textit{Care, Fairness, Loyalty, Authority, Sanctity}) and corresponding negative ethics (\textit{Harm, Cheating, Betrayal, Subversion, Degradation}). We follow the same pipeline to reconfigure the multi-agent system to classify the moral alignment of each related sentence. The pilot validation yielded a Cohen's Kappa of 0.67 and the agent consistency score reaches a Fleiss's Kappa of 0.989, confirming the effectiveness of the system.

Through this process, the moral alignment for each essay is represented as a 10-dimensional value vector: $\mathbf{m} = [m_1, \dots, m_{10}]$, where $m_i$ denotes the normalized frequency of a specific moral category within the text. Finally, we compute the moral alignment ($MA$) Shannon entropy to quantify the moral diversity of an essay:

 \vspace{-0.5em}

\begin{equation}
MA = -\sum_{i=1}^{10} \frac{m_i}{n} \log \frac{m_i}{n}
\end{equation}

A higher $MA$ score signifies that the student has engaged with a diverse set of moral perspectives.

%% file: 3.3.adoption.tex
\subsection{Feedback Uptake Quantification}

To unfold how students incorporate the feedback from the hybrid LLM-teacher system into their revision, we develop a four-step pipeline as follows:




\paragraph{Suggestion Extraction and Classification.}

For each essay, we collect the raw LLM-generated suggestions $\mathcal{S}_{\text{initial}}$ from stage 2 and the instructor-mediated finalized suggestions $\mathcal{S}_{\text{final}}$ from stage 3, and then map all suggestions into a shared semantic space using Sentence-BERT (SBERT).

For each finalized suggestion $s_j \in \mathcal{S}_{\text{final}}$, we calculate its maximum similarity against the raw LLM set $s_i \in \mathcal{S}_{\text{initial}}$: $\max_{s_i \in \mathcal{S}_{\text{initial}}} (\cos(v_j, v_i))$ where $v_i$ and $v_j$ are the vector embeddings of $s_i$ and $s_j$. If it is larger than the threshold $\delta_{m}=0.75$, the suggestion is categorized as $\mathcal{S}_{\text{L}}$ representing teacher retention of LLM feedback. Otherwise, it is categorized as $\mathcal{S}_{\text{T}}$, meaning the teacher significantly revises the LLM output or injects novel contents.

Finally, we categorize each suggestion $s_j \in \mathcal{S}_{\text{final}}$ based on its fine-grained linguistic focus into one of six dimensions. We provide the categorization method in Appendix B.

\paragraph{Revision Candidate Identification}
To isolate the student's response to feedback, we identify the subset of sentences in the revised draft $D_{post}$ that underwent meaningful modification. Using the \texttt{difflib} sequence matching algorithm, we compute the similarity between sentences in $D_{pre}$ and $D_{post}$. We define a set of revision candidates $\mathcal{C} = \{c_1, c_2, \dots, c_m\}$, where a sentence $s \in D_{post}$ is included in $\mathcal{C}$ if its maximum similarity to any sentence in $D_{pre}$ $\max_{s' \in D_{pre}} \text{sim}(s, s')$ is lower than the threshold $\delta_{r} = 0.95$. This ensures that $\mathcal{C}$ represents only the active revisions made by the student in the second iteration while excluding those remain unchanged.



\paragraph{Attention-based Semantic Matching}
We develop an attention mechanism to determine if a suggestion $s_j \in \mathcal{S}_{\text{final}}$ is adopted by the student. In particular, given a suggestion $s_j$ and the revision candidates $\mathcal{C}$, we compute attention weights $\alpha_{ij}$ using a softmax function:

 \vspace{-1em}
\begin{equation}
    \alpha_{ji} = \frac{\exp(cos(v_j, v_i) / \tau)}{\sum_{i=1}^{m} \exp(cos(v_j, v_i) / \tau)}
\end{equation}
\noindent where $v_j$ and $v_i$ are the vector embeddings of $s_j$ in the suggestion set and $c_i$ in the revision candidate set, while $\tau=0.1$ is the temperature parameter. The matching score $Match_j$ for suggestion $s_j$ is the weighted sum of similarities:

 \vspace{-1em}

\begin{equation}
    Match_j = \sum_{i=1}^{m} \alpha_{ji} \cdot cos(v_j, v_i)
\end{equation}
 \vspace{-0.5em}

A finalized suggestion $s_j \in \mathcal{S}_{\text{final}}$ is considered adopted if its matching score $Match_j$ exceeds a specific adoption threshold $\delta_{a}=0.5$. 

\paragraph{Feedback Uptake Ratio Calculation}




The Feedback Uptake Rate (FUR) is the proportion of the total count of suggestions adopted by the student, Feedback Uptake Amount (FUA), relative to the total number of provided suggestions $|\mathcal{S}_{\text{final}}|$:

 \vspace{-1.5em}
\begin{equation}
\text{FUA} = \sum_{j=1}^{|\mathcal{S}_{\text{final}}|} \mathbb{I}(\text{Match}_{j} > \delta_{a})
\quad
\text{FUR} = \frac{\text{FUA}}{|\mathcal{S}_{\text{final}}|}
\end{equation}

 \vspace{-0.5em}

\noindent where $\mathbb{I}(\cdot)$ denotes the indicator function. To evaluate the specific impact of teacher intervention, we decompose these metrics into $\text{FUA}_L$ and $\text{FUR}_L$ for suggestions from the LLM subset ($\mathcal{S}_L$), and $\text{FUA}_T$ and $\text{FUR}_T$ for those refined by the teacher ($\mathcal{S}_T$). 


%% file: 4.experiments-and-results.tex
\section{Experiments and Results}

\subsection{System Implementation and Dataset}

The hybrid LLM-teacher system was implemented and deployed in a real-world K-12 writing platform starting in 2023. For this study, we focus on the process where a student submitted a revised draft  following the receipt of the feedback, resulting in a large-scale dataset centered on this iterative revision process. To ensure data integrity, we excluded invalid submissions lacking corresponding LLM or teacher grades. Then, all scores were normalized to a standard $[0, 100]$ scale. Furthermore, we applied a 2.5\% trimming strategy based on sentence and token counts to remove outliers. 

As reported in Table \ref{tab:dataset_stats}, the dataset spanned over two years, involving $10,195$ students across $120$ distinct schools. The resulting corpus comprises $57,954$ essays derived from $1,602$ unique writing tasks, yielding a total of $28,977$ student-task pairs. On average, each essay contains $129.01$ tokens and $14.56$ sentences. Each pair captures a complete cycle of the  writing process, consisting of the initial draft ($D_{pre}$), the raw LLM-generated suggestion ($\mathcal{S}_{\text{initial}}$), the instructor-mediated finalized suggestion ($\mathcal{S}_{\text{final}}$) and the revised version ($D_{post}$). 

\begin{table}[t]
\centering
\small
\setlength{\tabcolsep}{4pt} 
\caption{Statistics of the constructed dataset.}
 \vspace{-0.5em}

\begin{tabular}{l|r}
\toprule
\textbf{Statistic} & \textbf{Value} \\
\midrule
Time Span & 2023.05 -- 2025.07 \\
\# Schools & 120 \\
\# Students & 10,195 \\
\# Writing Tasks & 1,602 \\
\midrule
\textbf{Essay} & \\
\# Total Essays & 57,954 \\
\# Average Sentences & 14.56 \\
\# Average Tokens & 129.01 \\
\# Student-Task Pairs & 28,977 \\
\midrule
\textbf{Grade (Mean)} & \\
$\text{grade}_{\text{LLM}}^{pre}$ vs $\text{grade}_{\text{LLM}}^{post}$  & 80.74 vs 85.75 $^{***}$\\
$\text{grade}_{\text{Teacher}}^{pre}$ vs $\text{grade}_{\text{Teacher}}^{post}$  &  75.17 vs 80.34 $^{***}$\\
\bottomrule
\end{tabular}
 \vspace{-2.0em}
\label{tab:dataset_stats}
\end{table}

Notably, on average, \textbf{student's performance significantly\footnote{Given the non-normal distribution of the data (Shapiro-Wilk test, $p < 0.001$), we employ the Wilcoxon Signed-Rank Test for paired comparisons and the Mann-Whitney U Test for independent samples. In this study, statistical significance levels are denoted as follows: $^{***}:p < 0.001;^{**}:p < 0.01;^{*}:p < 0.05;^{\dagger}:p < 0.1;{^{ns}}:p > 0.1;$} improved across both evaluative lenses}: teacher-assigned grades increased by $5.17$ points, from $\text{grade}_{\text{Teacher}}^{pre}$ to $\text{grade}_{\text{Teacher}}^{post}$, and LLM-assigned grades by $5.01$ points, from $\text{grade}_{\text{LLM}}^{pre}$ to $\text{grade}_{\text{LLM}}^{post}$ in the revised versions. Notably, the increase in teacher-assigned grades was significantly greater than the increase in LLM-assigned grades. These confirm the efficacy of our triadic system in fostering measurable writing skill improvement.


\subsection{Results}

\input{4.3.1.finding}

\input{4.3.2.finding}
\input{4.3.3.finding}

%% file: 4.3.1.finding.tex
\subsubsection{Multidimensional Linguistic Growth}

\paragraph{Holistic Writing Skill Growth cross ALL Linguistic Dimensions with Pronounced Gains in the Interpersonal Function.}
As detailed in Table \ref{tab:linguistic_shifts}, we observe a statistically significant improvement across all six dimensions, revealing a holistic writing skill improvement. Notably, the interpersonal function demonstrates the most substantial growth, with the emotional spectrum expanding by $7.719\%$ and moral alignment by $5.735\%$. This suggests that our system transcends mere formal correction: \textbf{beyond refining students' writing mechanics and structural cohesion, the hybrid LLM-teacher intervention can foster deeper ideological maturity and cognitive complexity}, enabling students to navigate more nuanced emotional and moral landscapes in their revisions.

 \vspace{-0.5em}

\begin{table}[ht]
\centering
\small
\caption{Linguistic growth between the initial drafts ($D_{pre}$) and the refined versions ($D_{post}$).} 
 \vspace{-0.5em}

\setlength{\tabcolsep}{3pt} 
\begin{tabular}{lccc}
\toprule
\textbf{Dimension} & $\text{Mean}_{\text{pre}}$ & $\text{Mean}_{\text{post}}$ & $\Delta\%$ \\
\midrule  
Lexical Richness & 0.850 & 0.875 & $+3.313^{***}$ \\
Syntactic Diversity & 0.696 & 0.701 & $+1.334^{***}$ \\
Semantic Dispersion & 0.483 & 0.486 & $+1.556^{***}$ \\
Semantic Shift & 0.524 & 0.528 & $+1.641^{***}$ \\
Emotional Spectrum & 1.115 & 1.231 & $+7.719^{***}$ \\
Moral Alignment & 0.480 & 0.536 & $+5.735^{***}$ \\
\bottomrule
\end{tabular}
 \vspace{-1.5em}

\label{tab:linguistic_shifts}
\end{table}

\paragraph{Shift Toward Pro-social Emotional and Moral Narratives.}
Table~\ref{tab:affective_moral_combined} reveals a distinct polarity shift for the interpersonal function improvement: in approach-oriented emotions, including \textit{Anticipation, Joy, Trust and Fear}, all exhibiting significant growth. However, avoidance-oriented emotions, specifically Anger and Disgust, are significantly reduced. A similar pro-social effect appears in moral alignment. The positive foundations, including \textit{Sanctity, Care, and Loyalty}, increase significantly, while negative foundations (\textit{Harm, Cheating, Subversion, and Degradation}) are consistently suppressed. Notably, decreases in \textit{Authority} and \textit{Fairness} suggest a shift toward more egalitarian narrative styles. \textbf{These indicate that the intervention steers students away from conflict-based toward more pro-social narratives.}

\begin{table}[ht]
\centering
\footnotesize 
\caption{Shifts in emotional and moral dimensions}
 \vspace{-0.5em}

\setlength{\tabcolsep}{2pt} 
\renewcommand{\arraystretch}{0.85} 
\begin{tabular}{lr|lr} 
\toprule
\textbf{Emotion} & \textbf{$\Delta\%$} & \textbf{Moral} & \textbf{$\Delta\%$} \\
\midrule
\multicolumn{2}{l|}{\textbf{I. Approach-oriented}} & \multicolumn{2}{l}{\textbf{I. Positive}} \\
Anticipation & $+2.498^{***}$ & Sanctity & $+1.513^{***}$ \\
Joy & $+1.789^{***}$ &  Care & $+0.767^{***}$\\
Trust & $+0.975^{***}$ &  Loyalty & $+0.509^{***}$\\
Fear & $+0.305^{***}$ &  Authority & $-0.127^{***}$\\
& &  Fairness & $-0.021^{**}$ \\
\multicolumn{2}{l|}{\textbf{II. Avoidance-oriented}} & \multicolumn{2}{l}{\textbf{II. Negative}} \\
Surprise & $+0.235^{***}$ &  Degradation & $-0.017^{**}$\\
Sadness & $+0.212^{ns}$ &  Harm & $-0.028^{***}$\\
Disgust & $-0.073^{***}$ &  Betrayal & $-0.003^{ns}$ \\
Anger & $-0.141^{***}$ &  Subversion & $-0.029^{***}$\\
& &  Cheating & $-0.052^{***}$\\
\bottomrule
\end{tabular}
 \vspace{-1em}

\label{tab:affective_moral_combined}
\end{table}

%% file: 4.3.2.finding.tex
\subsubsection{Labor Division and Feedback Adoption}

\paragraph{Hybridized Team Collaboration with Teacher's Pedagogical Bridge Role.} 
As reported in Table \ref{tab:combined_agent_analysis}, the collaborative synergy between LLM and teacher results in a hybridized feedback landscape where the final suggestion pool ($\mathcal{S}_{\text{final}}$) consists of 58.2\% raw LLM suggestions ($\mathcal{S}_{\text{L}}$) and 41.8\% teacher-mediated refinements ($\mathcal{S}_{\text{T}}$). The LLM provides the majority of the volume, particularly in the moral alignment category, which is almost entirely LLM-driven ($98.912\%$), while teachers provide a higher relative contribution to the semantic shift dimension to ensure logical cohesion. 

However, teachers demonstrate a significantly higher feedback adoption ratio overall and across nearly all individual dimensions, with the sole exception of non-significant difference in semantic shift. Notably, our design ensured that students could not distinguish the source of the provided feedback directly. This ``trust premium'' for teacher feedback confirms that the teacher's refinements, acting as a ``Pedagogical Bridge'', possess an inherent qualitative alignment with student needs that raw LLM output lacks. Hence, \textbf{while the LLM acts as the primary generative engine, teacher intervention remains the critical bridge for actual student implementation.}

\begin{table}[htbp]
\centering
\footnotesize
\caption{Finalized suggestions contributed by LLM and teacher, and their adoption rate}
 \vspace{-1em}

\setlength{\tabcolsep}{2pt} 
\setlength{\abovecaptionskip}{3pt}   
\setlength{\belowcaptionskip}{-6pt}  
\setlength{\tabcolsep}{1pt} 
\begin{tabular}{lcccccc}
\toprule
 & \multicolumn{3}{c}{\textbf{Feed. Share (\%)}} & \multicolumn{3}{c}{\textbf{FUR (\%)}} \\
\cmidrule(lr){2-4} \cmidrule(lr){5-7}
\textbf{Dimention} & $\mathcal{S}_{L}$ & $\mathcal{S}_{T}$ & \textbf{Sig.} & $\text{FUR}_{L}$ & $\text{FUR}_{T}$ & \textbf{Sig.} \\
\midrule
Lexical Richness    & 53.037 & 46.963 & $^{***}$ & 85.368 & 91.761 & $^{***}$ \\
Syntactic Diversity  & 62.847 & 37.153 & $^{***}$ & 79.369 & 93.539 & $^{***}$ \\
Semantic Dispersion & 76.280 & 23.720 & $^{***}$ & 84.573 & 91.311 & $^{***}$ \\
Semantic Shift & 39.140 & 60.860 & $^{***}$ & 92.621 & 93.701 & $^{ns}$     \\
Emotional Spectrum  & 54.075 & 45.925 & $^{***}$ & 88.926 & 94.026 & $^{***}$ \\
Moral Alignment     & 98.912 & 1.088  & $^{***}$ & 82.699 & 84.463 & $^{*}$    \\
\midrule
Overall    & 58.158 & 41.842 & $^{***}$ & 84.688 & 91.895 & $^{***}$ \\
\bottomrule
\end{tabular}
 \vspace{-1.5em}
\label{tab:combined_agent_analysis}
\end{table}


\paragraph{Strategic Labor Division with the LLM as Generative Engine and Teacher as Pedagogical Gatekeeper.} 

We quantify \textit{Teacher Effort} as the volume of modifications to LLM output plus manual additions, calculated as $|\mathcal{S}_{\textbf{initial}} - \mathcal{S}_{\textbf{L}}| + \mathcal{S}_{\textbf{T}}$. We distinguish between two scenarios: \textit{Creation}, where the teacher drafts feedback from scratch, and \textit{Modification}, where the teacher refines LLM-generated content.

\begin{table}[htbp]
\centering
\caption{Teacher effort under creation and modification}
\vspace{-0.5em}

\footnotesize 
\setlength{\tabcolsep}{2pt} 
\setlength{\abovecaptionskip}{3pt} 
\setlength{\belowcaptionskip}{-5pt}

\begin{tabular}{lccc} 
\toprule
\textbf{Dimension} & \textbf{Creation} & \textbf{Modification} & \textbf{Sig. ($\Delta$)} \\
\midrule
Lexical Richness         & 1.400 & 0.026 & $^{***}\downarrow$ \\
Syntactic Diversity      & 0.290 & 0.017 & $^{***}\downarrow$ \\
Semantic Dispersion      & 0.232 & 0.014 & $^{***}\downarrow$ \\
Semantic Shift           & 0.520 & 0.024 & $^{***}\downarrow$ \\
Emotional Spectrum       & 0.270 & 0.022 & $^{***}\downarrow$ \\
Moral Alignment          & 0.011 & 0.026 & $^{***}\uparrow$ \\
\midrule
Overall                  & 3.951 & 0.095 & $^{***}\downarrow$ \\
\bottomrule
\end{tabular}
 \vspace{-1em}

\label{tab:ai_feedback_comparison}
\end{table}

As reported in Table \ref{tab:ai_feedback_comparison}, \textbf{LLM integration significantly reduces the teacher's burden}. In the creation scenario, total teacher effort is $3.951$, dominated by lexical richness ($1.400$), followed by semantic shift ($0.520$) and syntactic diversity ($0.290$). In contrast, in the modification scenario, total teacher effort is remarkably low ($0.095$), \textbf{representing a nearly 40-fold decrease in workload}. Beyond lexical richness, interventions are redistributed to moral alignment ($0.026$) and the emotional spectrum ($0.024$).
Notably, when transitioning from creation to modification, moral alignment is the only dimension where teacher effort increases. This highlights the role of the teacher not just as an editor, but as a moral gatekeeper, ensuring that automated suggestions align with the specific sensitive requirements of K-12 educational standards. In other words, \textbf{the triadic collaboration does not simply replace the teacher, but reorients human expertise toward high-stakes qualitative gatekeeping, mitigating the instructor burnout common in K-12 writing programs.}

%% file: 4.3.3.finding.tex
\subsubsection{The Triadic Collaboration Impact}

\paragraph{Two-channel Feedback Efficacy with Linguistic Dimensional Mediation.}

As shown in Model 1 (Table \ref{tab:regression_final_single}), the volume of adopted suggestions from both the LLM ($\text{FUA}_L$) and the teacher ($\text{FUA}_T$) significantly predicts improvements in writing quality. \textbf{This confirms the baseline efficacy of both channels, LLMs and teachers, in the revision process. Notably, teacher-mediated suggestions yield a higher marginal coefficient than raw LLM suggestions.} This suggests that human intervention effectively ``aligns'' student compositions with high-level evaluative criteria, confirming the teacher's role as a linguistic anchor that grounds generative output in pedagogical reality.

Model 2 further elucidates how specific SFL dimensions mediate these gains. Both the interpersonal metafunction (encompassing emotional spectrum and moral alignment) and the ideational metafunction (specifically lexical richness) emerge as potent positive mediators. While syntactic diversity contributes positively to LLM evaluations, it demonstrates a non-significant impact on final teacher grades, suggesting a discrepancy between algorithmic fluency and human-perceived quality. A critical evaluative divergence is observed within the textual metafunction: while semantic shift (local logical flow) contributes positively to the grade, semantic dispersion (global thematic breadth) actually penalizes the final score. As demonstrated in our functional division of labor analysis, LLMs intend to prioritize semantic dispersion whereas teachers prioritize semantic shift, suggesting \textbf{teacher's crucial role in mitigating LLM's potential over-expansion}.

\begin{table}[htbp]
\centering
\footnotesize
\caption{Fixed effects regression results. All models include student, teacher, and task fixed effects. Standard errors are reported in parentheses.}
\vspace{-0.5em}

\label{tab:regression_final_single}
\setlength{\tabcolsep}{0pt}
\begin{tabular*}{\columnwidth}{@{\extracolsep{\fill}}lcccc}
\toprule
 & \multicolumn{2}{c}{\textbf{Model 1}} & \multicolumn{2}{c}{\textbf{Model 2}} \\
\cmidrule(lr){2-3} \cmidrule(lr){4-5}
\textbf{Predictors} & \textbf{Tchr} & \textbf{LLM} & \textbf{Tchr} & \textbf{LLM} \\ 
\midrule
$\text{FUA}_L$ & 0.005** & 0.008*** & 0.004* & 0.007*** \\
 & (0.002) & (0.001) & (0.002) & (0.001) \\
$\text{FUA}_T$ & 0.005*** & 0.011*** & 0.005*** & 0.010*** \\
 & (0.001) & (0.001) & (0.001) & (0.002) \\ 
\midrule
Lexical Richness & -- & -- & 0.003*** & 0.008*** \\
 & & & (0.001) & (0.001) \\
Syntactic Diversity & -- & -- & -0.001 & 0.002* \\
 & & & (0.001) & (0.001) \\
Semantic Dispersion & -- & -- & -0.001 & -0.004** \\
 & & & (0.001) & (0.001) \\
Semantic Shift & -- & -- & 0.002$\dagger$ & 0.004*** \\
 & & & (0.001) & (0.001) \\
Emotion Spectrum & -- & -- & 0.003*** & 0.008*** \\
 & & & (0.001) & (0.001) \\
Moral Alignment & -- & -- & 0.003*** & 0.007*** \\
 & & & (0.001) & (0.001) \\ 
\midrule
Baseline Score & -0.009*** & -0.007*** & -0.009*** & -0.006*** \\
 & (0.000) & (0.000) & (0.000) & (0.000) \\ 
\midrule
\textbf{R-squared} & 0.880 & 0.566 & 0.880 & 0.578 \\
\textbf{Observations} & 23,647 & 23,647 & 23,467 & 23,467 \\ 
\bottomrule
\end{tabular*}
\vspace{-1.5em}
\end{table}


\begin{figure}[htbp]
  \centering
  \begin{subfigure}{0.45\columnwidth}
    \centering
    \includegraphics[width=\linewidth]{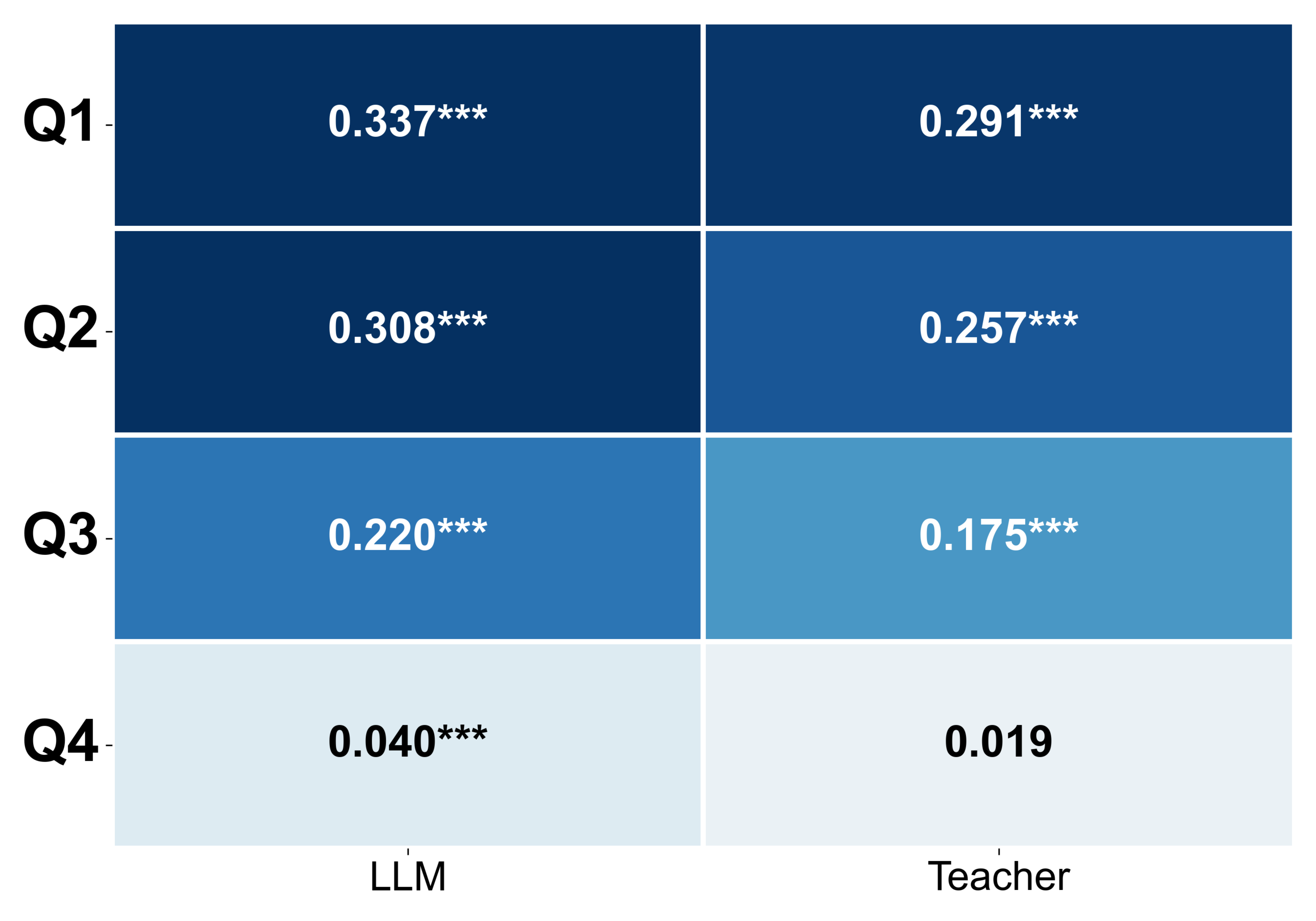}
    \caption{\small Lexical Richness}
  \end{subfigure}
  \hspace{0.1em} 
  \begin{subfigure}{0.45\columnwidth}
    \centering
    \includegraphics[width=\linewidth]{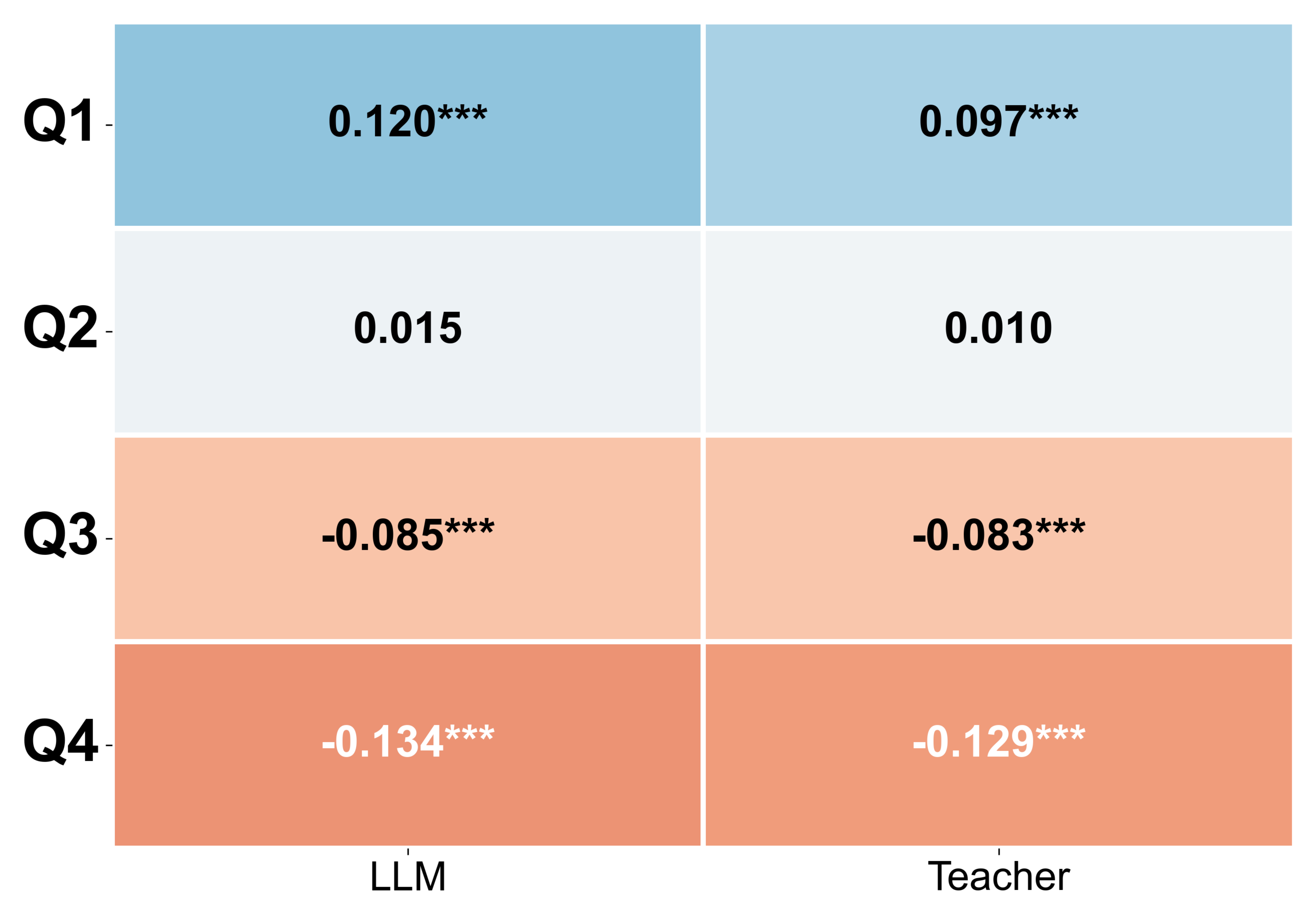}
    \caption{\small Syntactic Diversity}
  \end{subfigure}
  
  \vspace{0.5em}
  
  \begin{subfigure}{0.45\columnwidth}
    \centering
    \includegraphics[width=\linewidth]{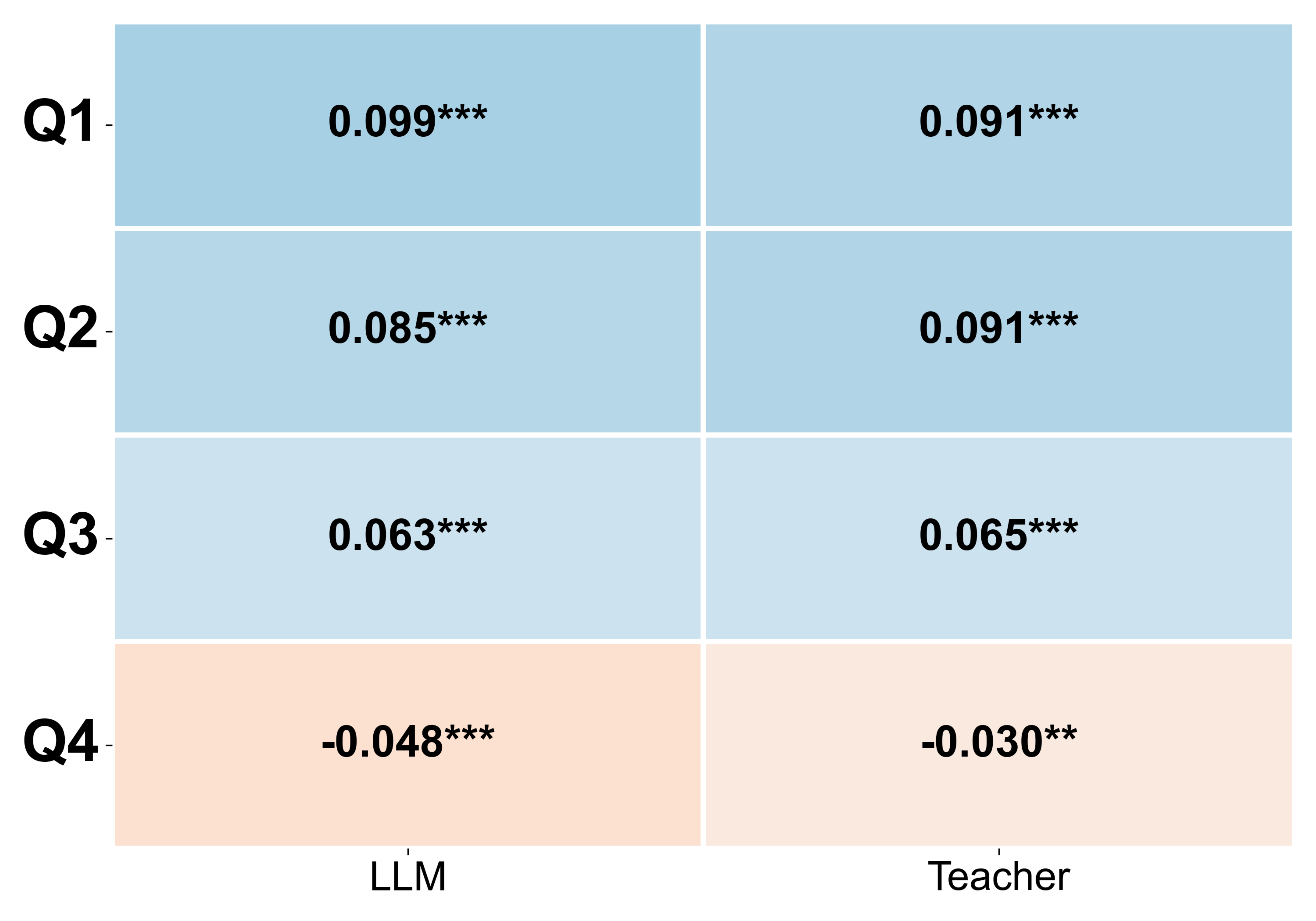}
    \caption{\small Semantic Dispersion}
  \end{subfigure}
  \hspace{0.1em}
  \begin{subfigure}{0.45\columnwidth}
    \centering
    \includegraphics[width=\linewidth]{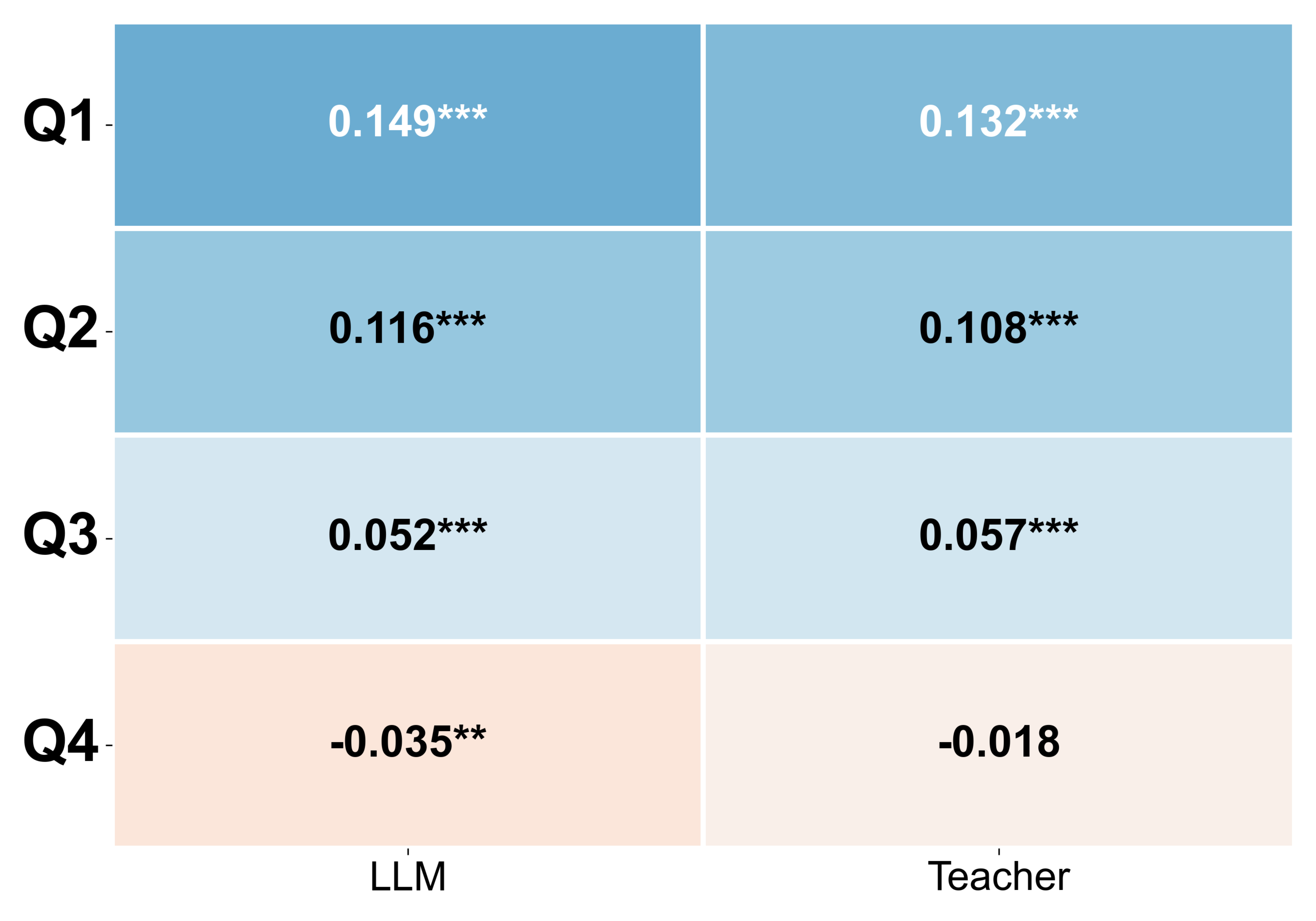}
    \caption{\small Semantic Shift}
  \end{subfigure}
  
  \vspace{0.5em}
  
  \begin{subfigure}{0.45\columnwidth}
    \centering
    \includegraphics[width=\linewidth]{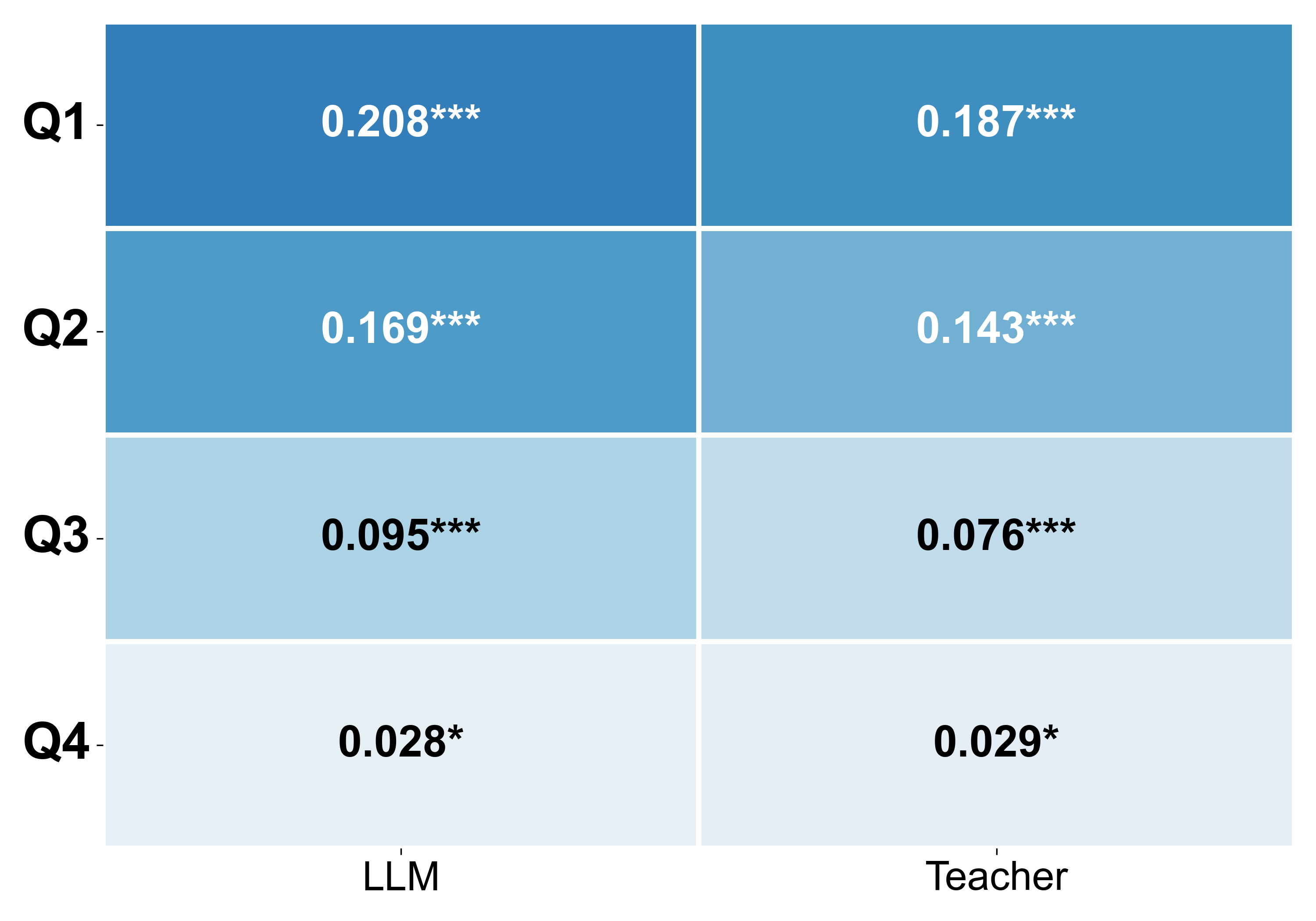}
    \caption{\small Emotional Spectrum}
  \end{subfigure}
  \hspace{0.1em}
  \begin{subfigure}{0.45\columnwidth}
    \centering
    \includegraphics[width=\linewidth]{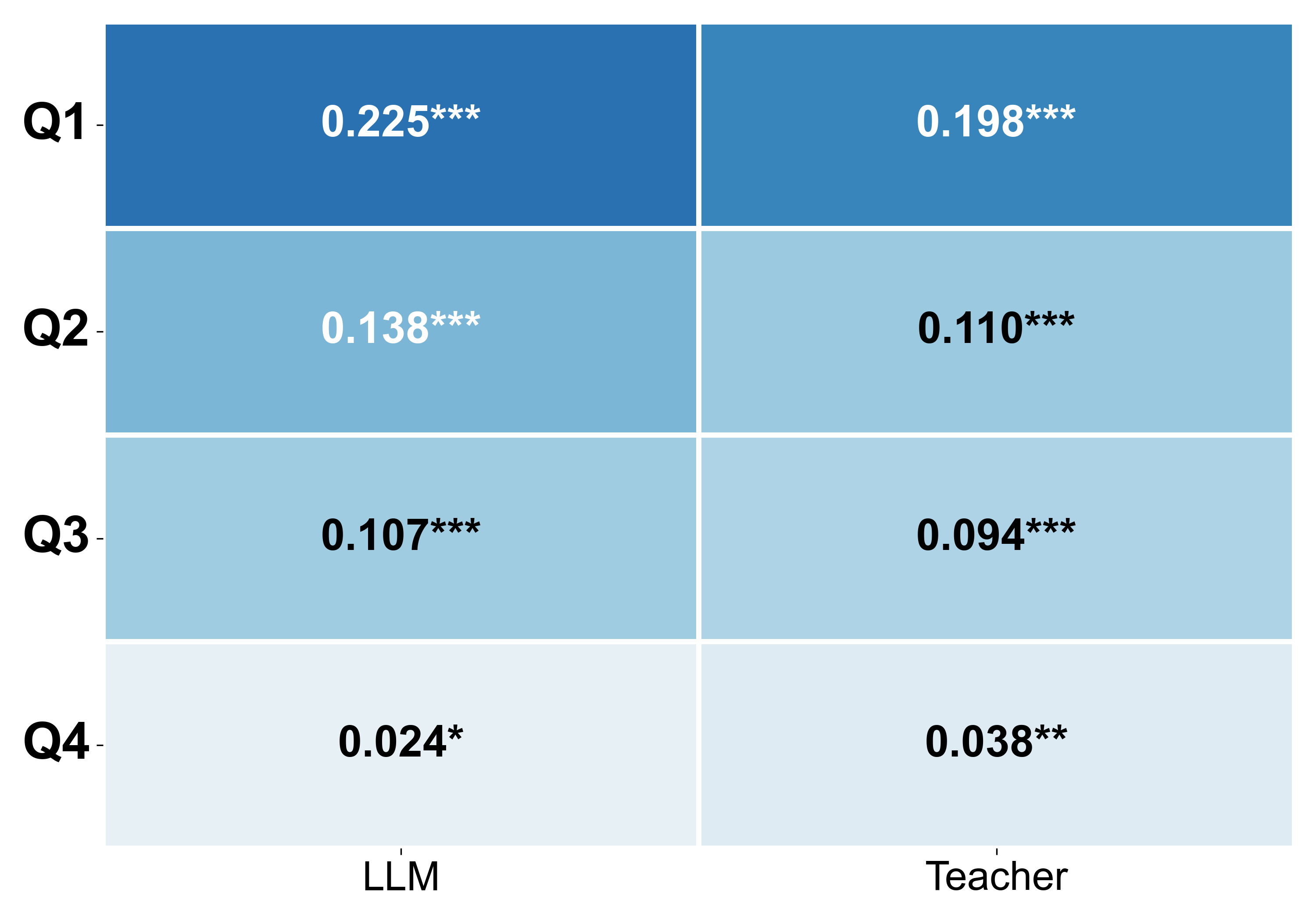}
    \caption{\small Moral Alignment}
  \end{subfigure}
  
  \vspace{-0.5em}
  \caption{Spearman correlation between linguistics growth and grade split into baseline quartiles.}
  \vspace{-1.5em}
  \label{fig:rq3_heatmaps_col}
\end{figure}

\paragraph{Diminishing Marginal Utility and the Proficiency-Based Ceiling Effect.} 
To unfold the mediation contribution from the linguistic growth, we analyze Spearman correlations between dimensional gains and grade improvements by partitioning the sample into quartiles based on initial linguistic proficiency. 

As illustrated in Figure \ref{fig:rq3_heatmaps_col}, a consistent pattern emerges across the proficiency spectrum: correlation coefficients steadily decrease from the lower to the higher quartiles. For dimensions such as syntactic diversity, semantic dispersion, and semantic shift, these correlations eventually turn negative in the highest quartiles ($Q4$). This trend indicates that while \textbf{lower-performing students derive significant pedagogical value from linguistic expansion, higher-performing students encounter a pronounced ``ceiling effect''}:
further increases in structural or thematic complexity may no longer translate into higher scores which can even become counterproductive. This suggests a dynamic, proficiency-aware triadic system that adjusts its strategy (i.e. adjusting LLM's system prompts) according to student's linguistic proficiency.

%% file: 5.conclusion.tex
\section{Conclusion}

This study implements and deploys a triadic collaboration model, integrating LLMs, teachers and students, within real-world K-12 writing environment. By conducting the first large-scale empirical dataset regarding the iterative process and quantifying the linguistic skill improvement from this collaboration, this study provides the empirical evidence demonstrating the efficacy of the designed triadic collaboration model. 

More importantly, the system achieves an effective labor division: the LLM functions as a highly efficient generative engine to mitigate instructor burnout while teacher intervention shifts to a strategic gatekeeper and bridge for suggestion adoption. Moreover, while both LLM and teacher contribute to writing skill improvement, we identify a proficiency-based ceiling effect: while linguistic expansion significantly aids novice writers, higher-proficiency students encounter diminishing marginal utility where excessive complexity can become counterproductive. These findings advocate for a triadic collaboration system making LLMs sharp tools for K-12 education.

%% file: 6.limitation.tex
\section*{Limitations}

While this study provides empirical evidence for the triadic collaboration model when implementing LLMs for K-12 students, several limitations warrant consideration:

First, the generalizability of the SFL framework used to quantify linguistic growth may vary across writing genres. While the SFL framework has been widely adopted in linguistic evaluation, the weight of specific linguistic dimensions might manifest differently. Exploring how the impact of these dimensions varies across genres (e.g., argumentative, expository, narrative, analytical) would be highly valuable for refining AI feedback strategies.

Second, this research focuses on the short-term impact of our triadic collaboration model. We did not fully consider the long-term longitudinal effects, such as the potential emergence of automation bias. For example, teachers might become increasingly over-reliant on the LLM's generative engine, potentially degrading the ``strategic gatekeeping'' and ``pedagogical bridge'' roles we identified. 

Additionally, our study is bounded by the specific capabilities of the underlying LLMs. In real-world K-12 environments, it is often technically or financially unfeasible to frequently update or switch underlying models. Future work should investigate the implementation of the suggested dynamically adaptive triadic system to test its efficacy and robustness. 

Finally, several external factors (such as the AI literacy of teachers and the socio-economic diversity of school environments) may influence the results. While we employ the fixed effect to control these factors' impact, assessing the equity of our system across these variables would be highly valuable.

%% file: 7.ethical.tex
\section*{Ethical Consideration and Design}

The responsible deployment of Large Language Models (LLMs) in K-12 educational settings is paramount. We identify and address three primary ethical domains critical to our triadic collaboration model:

First, a central ethical concern in educational NLP is the potential for LLMs to displace human educators. Our study provides strong empirical evidence against this ``automation'' narrative by demonstrating that the triadic model redefines, rather than removes, the teacher's role. By offloading repetitive linguistic drafting to the LLM (the "Generative Engine"), the system creates space for teachers to focus on high-stakes pedagogical gatekeeping and bridging. Our results show that both teacher and LLM interventions are indispensable for student development; however, teachers maintain a higher adoption rate and a higher marginal coefficient for improvement. This confirms that LLM implementation must be designed to mitigate teacher burnout while maintaining human accountability at the center of the learning process.

Second, to prevent cognitive disengagement, our model avoids generating ``raw'' complete writing. Instead, the LLM acts as a teaching assistant, providing granular suggestions that require student evaluation and revision. The teacher acts as a pedagogical bridge, filtering these suggestions to ensure alignment with specific educational objectives. Furthermore, our discovery of the proficiency-based ceiling effect serves as an ethical safeguard: it warns against the risks of ``over-scaffolding'' advanced learners with excessive linguistic complexity. By advocating for a dynamically adaptive system, we protect student agency and ensure that LLM intervention supports genuine skill acquisition rather than mere automated completion.

Third, this study was conducted with the approval of participating school administrations and was deemed exempt from Institutional Review Board (IRB) review as it did not involve direct intervention on students. Nevertheless, we strictly adhered to standard IRB ethics and privacy requirements. We utilized exclusively de-identified textual data, ensuring no Personally Identifiable Information (PII) was processed. All analysis was conducted within a secure, access-restricted environment, effectively controlling for any potential risks of data leakage.

\section*{Declaration of Generative AI in the Writing Process.}

During the preparation of this work the authors used LLMs like ChatGPT, Gemini, Claude and Deepseek, in order to improve language and readability. After using the generative AI service, the authors reviewed and edited the content and take full responsibility for the content of the submission.

%% file: appendix_annotation.tex
\begin{CJK}{UTF8}{gbsn}

\section{Automatic Emotion and Moral Annotation Pipeline}
\begin{figure}[H]
    \centering
    \includegraphics[width=0.95\linewidth]{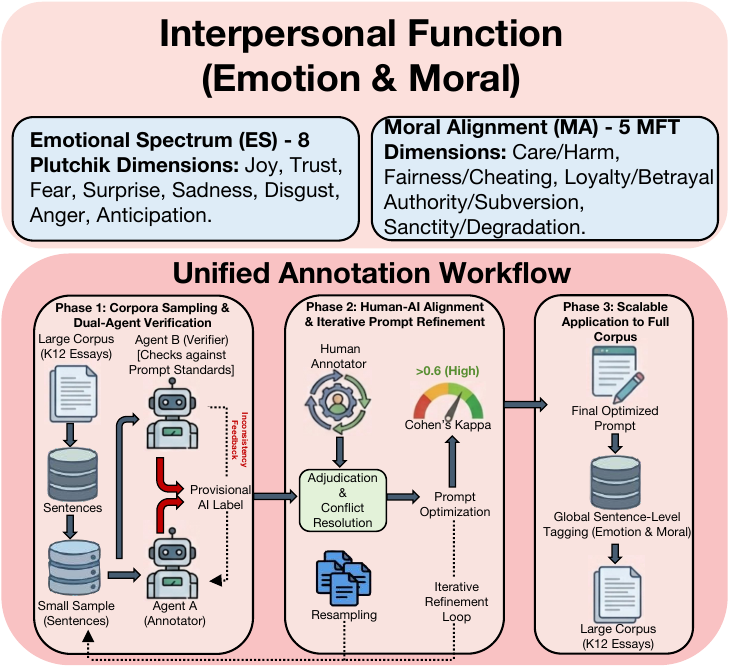}
    \caption{
        \textbf{Schematic overview of the automatic emotion and moral annotation pipeline.}
        Essays are segmented into sentences and filtered for emotion or moral relevance.
        Relevant sentences are annotated using an iterative LLM-based framework with critical feedback (A--B--A).
        Annotation prompts are calibrated based on human--model agreement before large-scale annotation.
        Sentence-level labels are finally aggregated at the writing level for distributional analysis.
    }
    \label{fig:annotation_pipeline}
\end{figure}

Figure~\ref{fig:annotation_pipeline} provides an overview of the complete annotation workflow. We describe each stage in detail below. In this study, both emotion and moral labels are generated through an automated annotation pipeline designed for large-scale text processing. The pipeline consists of four main stages: sentence-level preprocessing and relevance filtering; an iterative large language model (LLM) annotation framework guided by critical feedback (A--B--A); prompt calibration based on human--model agreement; and large-scale full annotation. To ensure the usability and reproducibility of the full annotation results, the full-run stage incorporates strict output format validation and completeness control. Missing samples are eliminated through set-difference detection and repeated re-runs until coverage is complete. Finally, sentence-level outputs are aggregated at the writing level to produce vectorized indicators that reflect the content structure of each entire essay.

\subsection{Sentence-Level Preprocessing and Relevance Filtering}

The annotation pipeline starts with essay-level texts. The raw data are organized at the essay level, and each essay is assigned a stable identifier (writing\_id). This identifier is kept consistent across all intermediate files throughout the pipeline to enable cross-step alignment and traceability.
Given each writing\_id and its corresponding text, the essays are segmented into
sentence-level analysis units using Chinese punctuation marks
(e.g., ``。'', ``？'', ``！'')
and newline characters (\texttt{\textbackslash n}).
Within each essay, sentences are assigned a primary sentence-level key 
(sentence\_id) according to their order of appearance in the original text.

Before entering LLM-based relevance judgment, a rule-based coarse filtering step is applied to remove fragments that clearly lack semantic content, such as symbol-only strings or extremely short texts of no more than four Chinese characters. Subsequently, a large language model (GPT-4) is used to determine whether the remaining sentences are related to emotion or moral content. This stage outputs only a binary decision, classifying sentences into filter pass and filter reject (AI-based). Only sentences that pass the filter proceed to fine-grained classification.

\subsection{Iterative Sentence-Level Classification with Critical Feedback Guidance}

To improve the stability and consistency of automatic annotations, we adopt an iterative LLM-based annotation framework guided by critical feedback, structured as an A--B--A process. The workflow is as follows.

First, the primary annotation model (Agent A; GPT-4.1, temperature = 0) generates an initial classification result (A1), including the predicted label, a confidence score, and a brief justification. Next, a review model (Agent B; Claude Sonnet 4, temperature = 0) evaluates A1 and produces an agree/disagree judgment, a critical critique, and a suggested label. Finally, Agent A revises its initial decision based on the reviewer’s feedback and outputs the final classification result (A2), along with a final confidence score and justification.

The sentence-level final decision is taken from the A2 output. The same A--B--A mechanism and validation framework are used for both emotion and moral annotation tasks, with task-specific label sets fixed as Plutchik’s eight basic emotions for emotion classification and a 10-class Moral Foundations Theory label set for moral classification.
\subsection{Iterative Annotation with Critical Feedback and Large-Scale Full Annotation}

\subsubsection{Prompt Calibration Based on Human--Model Agreement}

Under the A--B--A sentence-level classification framework, model judgments are highly dependent on
explicit constraints in the prompts regarding task definitions, label semantics, and output formats.
To ensure that model classification logic aligns statistically with human understanding before
large-scale deployment, we conduct systematic iterative prompt calibration for both emotion and
moral classification tasks.

In each calibration round, 20 sentences are randomly sampled from the relevance-filtered sentence
pool to form a batch. Human annotators first label these sentences with emotion or moral categories.
In parallel, using the current version of the classification prompts, Agent~A and Agent~B perform a
single-round automatic classification on the same batch. Model outputs are then aligned sentence by
sentence with human annotations, and Cohen's Kappa is computed to measure human--model agreement.

Based on the observed agreement levels and error patterns in each round, prompts are revised and
refined before entering the next iteration. Calibration is conducted independently for the emotion
and moral tasks. The process terminates when the average Cohen's Kappa across three consecutive
iterations exceeds 0.6, indicating substantial consistency between system and human annotations.

Final agreement scores reach 0.61 for the emotion task and 0.67 for the moral task. The corresponding
prompts are then fixed as the unified configuration for subsequent large-scale automatic annotation.
For transparency and reproducibility, we report the complete system and task prompts for both emotion
and moral annotation below.

\small
\clearpage
\onecolumn
\setlength{\LTpre}{0pt}
\setlength{\LTpost}{0.8em}
\renewcommand{\arraystretch}{1.12}

\begin{longtable}{p{0.18\linewidth} p{0.78\linewidth}}
\caption{Emotion annotation prompts used in the A--B--A framework}
\label{tab:emotion-prompts}\\
\toprule
\textbf{Prompt Type} & \textbf{Content} \\
\midrule
\endfirsthead

\toprule
\textbf{Prompt Type} & \textbf{Content} \\
\midrule
\endhead

\midrule
\multicolumn{2}{r}{\emph{(Continued on next page)}} \\
\endfoot

\bottomrule
\endlastfoot

Agent A1 (task) &
\textbf{Task Description}\\
& Read each sentence carefully and identify the single most dominant emotion expressed.\\
& \textbf{Emotion definitions and label examples are provided in Table~\ref{tab:emotion-defs}.}\\[0.4em]
& \textbf{Decision Rules}\\
& -- Select exactly one emotion label.\\
& -- Output a confidence score in $[0,1]$, rounded to two decimals.\\
& -- Provide a brief reason (1--3 sentences) grounded in explicit textual cues.\\[0.4em]
& \textbf{Output Example (TSV)}\\
& \texttt{sentence\_id<TAB>emotion<TAB>confidence<TAB>reason}\\
& \texttt{S001<TAB>anger<TAB>0.80<TAB>The sentence contains explicit blame and irritation cues, indicating anger.}\\
\addlinespace[1.2em]

Agent B (critic) &
\textbf{Task Description}\\
& Review Agent~A1’s initial judgment and assess whether it is reasonable. Identify possible confusion and recommend a more appropriate emotion if needed.\\
& \textbf{Emotion definitions and label examples are provided in Table~\ref{tab:emotion-defs}.}\\[0.4em]
& \textbf{Decision Rules}\\
& -- Output \texttt{y} if you agree with Agent~A1, otherwise \texttt{n}.\\
& -- Provide a short critique (1--3 sentences).\\
& -- Recommend exactly one emotion label.\\[0.4em]
& \textbf{Output Example (TSV)}\\
& \texttt{sentence\_id<TAB>agree<TAB>critique<TAB>suggested\_emotion}\\
& \texttt{S001<TAB>n<TAB>The sentence emphasizes uncertainty and threat rather than confrontation.<TAB>fear}\\
\addlinespace[1.2em]

Agent A2 (final decision) &
\textbf{Decision Task}\\
& Determine the final emotion label after considering the critic’s feedback.\\
& \textbf{Emotion definitions and label examples are provided in Table~\ref{tab:emotion-defs}.}\\[0.4em]
& \textbf{Decision Rules}\\
& -- Output exactly one final emotion label.\\
& -- Output a final confidence score in $[0,1]$, rounded to two decimals.\\
& -- Explain in 1--3 sentences why the initial judgment is kept or revised.\\[0.4em]
& \textbf{Output Example (TSV)}\\
& \texttt{sentence\_id<TAB>final\_emotion<TAB>final\_confidence<TAB>final\_reason}\\
& \texttt{S001<TAB>fear<TAB>0.74<TAB>The sentence highlights uncertainty and perceived threat, aligning more with fear.}\\

\end{longtable}

\clearpage

\begin{longtable}{p{0.18\linewidth} p{0.78\linewidth}}
\caption{Emotion label definitions and examples (Plutchik-8)}
\label{tab:emotion-defs}\\
\toprule
\textbf{Label} & \textbf{Definition / Typical Cues} \\
\midrule
\endfirsthead

\toprule
\textbf{Label} & \textbf{Definition / Typical Cues} \\
\midrule
\endhead

\midrule
\multicolumn{2}{r}{\emph{(Continued on next page)}} \\
\endfoot

\bottomrule
\endlastfoot

anger &
Core meaning: Anger is an antagonistic emotional state that arises when an individual perceives that their goals are being obstructed, their rights are being violated, or they are being treated unfairly. Its core lies in blame toward an external target, hostility, or an impulse to retaliate. This emotion is typically accompanied by a clear attribution of responsibility, emphasizing conflict, opposition, and a motivation to change the status quo. It may manifest as overt indignation or exist in an implicit form such as dissatisfaction or suppression.\\
& Typical cues: getting angry, anger, complaining, accusing, dissatisfaction......\\
\addlinespace[0.9em]

anticipation &
Core meaning: Anticipation is a future-oriented emotional state, reflecting an individual’s psychological preparation and anticipatory attention toward upcoming events, potential outcomes, or consequences of actions. Its core is not an immediate experience of pleasure or pain, but a cognitive orientation toward future possibilities and motivational activation. It is often associated with planning, hope, vigilance, or moral orientation, and plays an important role in learning, growth, and the formation of goals.\\
& Typical cues: anticipation, looking forward to, hope, preparation, planning......\\
\addlinespace[0.9em]

disgust &
Core meaning: Disgust is an emotional reaction that stems from a strong sense of repulsion toward certain objects, behaviors, or situations. Its function is to prompt individuals to distance themselves from things perceived as harmful, unclean, or violating moral norms. This emotion can be grounded in physiological discomfort, and it can also extend to social and moral aversion and contempt, emphasizing a tendency to reject and sever contact.\\
& Typical cues: nausea, aversion, dislike, being despised......\\
\addlinespace[0.9em]

fear &
Core meaning: Fear is an emotional response that arises when an individual perceives threats, uncertainty, or potential danger. Its core lies in the anticipation of negative consequences and the resulting defensive psychological state. This emotion is typically accompanied by tension, anxiety, and vigilance, emphasizing a motivation to avoid risk or reduce losses.\\
& Typical cues: fear, worry, tension, unease......\\
\addlinespace[0.9em]

joy &
Core meaning: Joy is an emotional state triggered by satisfaction, success, or positive experiences, reflecting an individual’s positive evaluation of the current situation and pleasant feelings. Its core lies in the immediate emotional experience itself, often expressed as relaxation, delight, or happiness, and accompanied by acceptance and enjoyment of the current state, rather than being primarily characterized by future expectations or risk assessment.\\
& Typical cues: happiness, joy, being glad, satisfaction, excitement, enjoyment, warmth......\\
\addlinespace[0.9em]

sadness &
Core meaning: Sadness is a low emotional state that arises when an individual experiences loss, failure, setbacks, or powerlessness to change reality. Its core lies in an affective response to negative outcomes and inward-oriented regulation.\\
& Typical cues: sadness, sorrow, loss, depression, regret......\\
\addlinespace[0.9em]

surprise &
Core meaning: Surprise is an instantaneous emotional response triggered by unexpected events or information. Its core lies in a rapid cognitive reaction to broken expectations. This emotion itself does not have a stable positive or negative affective polarity; rather, as a transitional state, it prompts individuals to quickly adjust their cognitive framework to cope with a new situation.\\
& Typical cues: surprise, didn’t expect, actually, unexpectedly......\\
\addlinespace[0.9em]

trust &
Core meaning: Trust is an emotional experience based on the perceived reliability and safety of others, relationships, or the environment. Its core lies in reducing defensiveness, accepting dependence, and being willing to bear uncertainty in interactions. This emotion emphasizes stability and continuity, often accompanied by feelings of reassurance, identification, and reliance, and plays a role in social relationships and cooperative behavior.\\
& Typical cues: believe, feel assured, trust, reassurance......\\
\addlinespace[1.0em]
\multicolumn{2}{p{0.96\linewidth}}{%
\textbf{[Label examples]}\par
“My mother also understood how I felt at this moment, so she quietly walked out. I watched my mother’s departing back, and I suddenly felt a trace of guilt.”：sadness\par
“I still remember before—its leaves were almost withered, but today it actually bloomed. How amazing.”：surprise\par
“Dark clouds shrouded my heart; the light seemed out of reach.”：fear\par
“This kind of spirit is worth learning from.”：anticipation\par
“This time my essay score actually reached as high as ninety-three!”：surprise\par
“It felt as if I was pulled back into the long river of memories; when I was 7 years old, a warm recollection surged in my heart.”：joy\par
“From fairy tales to novels, the types of books have changed, but the feelings and inspiration they bring me have never changed.”：anticipation\par
“By reading \textit{Special Forces School}, I not only experienced the excitement and thrills of adventure, but also learned the importance of courage and wisdom when facing difficulties.”：anticipation\par
“We praised her for a long time.”：trust\par
“This is truly awful!”：disgust%
} \\

\end{longtable}

\clearpage

\setlength{\LTpre}{0pt}
\setlength{\LTpost}{0pt}
\renewcommand{\arraystretch}{1.10}

\vspace{1.5em}

\onecolumn

\small
\setlength{\LTpre}{0.6em}
\setlength{\LTpost}{0.6em}
\renewcommand{\arraystretch}{1.10}

\begin{longtable}{p{0.22\linewidth} p{0.74\linewidth}}
\caption{Moral annotation prompts used in the A--B--A framework}
\label{tab:moral-prompts}\\
\toprule
\textbf{Prompt Type} & \textbf{Content} \\
\midrule
\endfirsthead

\toprule
\textbf{Prompt Type} & \textbf{Content} \\
\midrule
\endhead

\midrule
\multicolumn{2}{r}{\emph{(Continued on next page)}} \\
\endfoot

\bottomrule
\endlastfoot

Agent A1 (task) &
\textbf{Task Description}\\
& Read each sentence carefully and identify the single most salient moral expressed in the text.\\
& \textbf{moral definitions, neutral-sentence rules, and reference examples are provided in Table~\ref{tab:moral-defs}.}\\[0.4em]
& \textbf{Decision Rules}\\
& -- Select exactly one moral label.\\
& -- Output a confidence score in $[0,1]$, rounded to two decimals.\\
& -- Provide a brief reason (1--3 sentences) grounded in explicit textual evidence.\\[0.4em]
& \textbf{Output Example (TSV)}\\
& \texttt{sentence\_id<TAB>moral<TAB>confidence<TAB>reason}\\
& \texttt{S001<TAB>care\_pos<TAB>0.83<TAB>The sentence explicitly evaluates helping others as morally appropriate, reflecting a care-oriented moral.}\\
\addlinespace[1.2em]

Agent B (critic) &
\textbf{Task Description}\\
& Review Agent~A1’s moral judgment and assess whether it is appropriate given the sentence focus. Identify potential confusion and recommend a more suitable moral if necessary.\\
& \textbf{moral definitions, decision standards, and examples are provided in Table~\ref{tab:moral-defs}.}\\[0.4em]
& \textbf{Decision Rules}\\
& -- Output \texttt{y} if you agree with Agent~A1, otherwise \texttt{n}.\\
& -- Provide a short critique (1--3 sentences).\\
& -- Recommend exactly one moral label.\\[0.4em]
& \textbf{Output Example (TSV)}\\
& \texttt{sentence\_id<TAB>agree<TAB>critique<TAB>suggested\_moral}\\
& \texttt{S001<TAB>n<TAB>The sentence emphasizes responsibility and rule adherence more than concern for others.<TAB>authority\_pos}\\
\addlinespace[1.2em]

Agent A2 (final decision) &
\textbf{Decision Task}\\
& Make the final moral assignment after considering the critic’s feedback.\\
& \textbf{moral definitions, neutral-sentence rules, and examples are provided in Table~\ref{tab:moral-defs}.}\\[0.4em]
& \textbf{Decision Rules}\\
& -- Output exactly one final moral label.\\
& -- Output a final confidence score in $[0,1]$, rounded to two decimals.\\
& -- Explain in 1--3 sentences why the initial judgment is retained or revised.\\[0.4em]
& \textbf{Output Example (TSV)}\\
& \texttt{sentence\_id<TAB>final\_moral<TAB>final\_confidence<TAB>final\_reason}\\
& \texttt{S001<TAB>authority\_pos<TAB>0.78<TAB>After considering the critique, rule-following and responsibility are the dominant evaluative focus.}\\

\end{longtable}
\clearpage
\begin{longtable}{p{0.18\linewidth} p{0.78\linewidth}}
\caption{Moral label definitions and guidelines (MFT-10)}
\label{tab:moral-defs}\\
\toprule
\textbf{Label} & \textbf{Definition / Typical Scenarios / Targets / Expressions} \\
\midrule
\endfirsthead

\toprule
\textbf{Label} & \textbf{Definition / Typical Scenarios / Targets / Expressions} \\
\midrule
\endhead

\midrule
\multicolumn{2}{r}{\emph{(Continued on next page)}} \\
\endfoot

\bottomrule
\endlastfoot

Care &
\textbf{Core definition:} Care refers to the significant emotional resonance and discomfort that arise when an individual observes others or animals in a state of suffering, together with a tendency to take action to reduce harm, provide help, and offer protection, thereby improving the target’s situation as much as possible. \par
\textbf{Typical triggering scenarios:} For example, when seeing someone who is ill but unattended, an elderly person with limited mobility, a neglected child, or a stray animal suffering from hunger and cold, people often naturally develop a judgmental tendency such as “Can I offer help?” or “How can their suffering be reduced?” \par
\textbf{Typical targets:} The targets of Care include not only humans but also animals; it may also be directed at “states requiring care” themselves, such as injury, hunger, cold, loneliness, and helplessness. \par
\textbf{Common expressions:} These include “That’s so pitiful,” “They shouldn’t have to endure this,” and “We should take care of them,” often followed by concrete supportive actions such as delivering medicine, providing companionship, offering food, reducing waste of resources, reducing environmental damage, or arranging care. \\
\addlinespace[0.8em]

Harm &
\textbf{Core definition:} Harm refers to inflicting physical or psychological damage on others (not limited to humans) despite knowing that the target may experience pain or discomfort, or showing indifference and disregard toward their suffering. \par
\textbf{Typical triggering scenarios:} For example, someone deliberately humiliates others with harsh language, mocks others in public to create embarrassment, deliberately frightens someone for amusement despite knowing their fear, abuses animals, or damages the environment. \par
\textbf{Typical targets:} The targets of Harm may be individuals or animals, as well as the dignity and psychological feelings of others (e.g., causing someone to lose face, feel fear, or feel wronged can also constitute Harm). \par
\textbf{Common expressions:} These include “It’s his fault for being so stupid,” “I just want to make him feel bad,” and “Let him cry, I don’t care,” or continuing to use harmful language under the pretext of joking, such as “I’m just kidding—why are you so sensitive,” as well as expressions involving “damaging the environment.” \\
\addlinespace[0.8em]

Fairness &
\textbf{Core definition:} Fairness refers to applying the same standards to different targets under the same circumstances, without granting differential treatment based on identity differences. \par
\textbf{Typical triggering scenarios:} For example, cutting in line while shopping is considered inappropriate because first-come-first-served should be followed; similarly, in competitions, awards, and admissions, relevant rules should apply equally to all participants. \par
\textbf{Typical targets:} The targets of Fairness may include individuals and groups, as well as the rules and procedures themselves, such as queueing rules, competition rules, selection procedures, and allocation mechanisms. \par
\textbf{Common expressions:} These include “Why does he get to cut in line,” “Everyone should follow the rules,” and “We should treat everyone equally,” emphasizing that “the same situations should be handled in the same way.” \\
\addlinespace[0.8em]

Cheating &
\textbf{Core definition:} Cheating refers to circumventing rules through covert means in the presence of explicit regulations in order to obtain unjust benefits, while attempting to appear as though nothing has happened. \par
\textbf{Typical triggering scenarios:} For example, peeking at answers or using cheat sheets during an exam; or deliberately tampering with an inaccurate scale or concealing expired goods in a transaction to avoid detection by others. \par
\textbf{Typical targets:} The targets of Cheating may include deceived individuals, as well as the exploited rules and transactional systems themselves, such as exam regulations, commercial integrity, or contractual agreements. \par
\textbf{Common expressions:} These include “No one will find out anyway,” “If I don’t do this, I’ll lose out,” and “Everyone does it,” with the core characteristics being “covert rule evasion, concern about being discovered, and gaining advantage through such actions.” \\
\addlinespace[0.8em]

Loyalty &
\textbf{Core definition:} Loyalty refers to treating “one’s own side” as in-group members and partners, being willing to stand by them and jointly bear responsibilities and risks, rather than easily abandoning or withdrawing. \par
\textbf{Typical triggering scenarios:} For example, being willing to support and share burdens when a family encounters hardship; or not leaving immediately when a team faces difficulties, but instead making efforts to overcome them together. \par
\textbf{Typical targets:} The targets of Loyalty may include family, friends, groups, or organizations, as well as “shared reputation and commitments,” such as “what we promised to complete together” or “this family/team must not fall apart.” \par
\textbf{Common expressions:} These include “Our own people should look out for each other,” “Since we agreed to do it together, we should do it together,” and “You can’t quit at the critical moment,” emphasizing “maintaining the in-group and not leaving lightly.” \\
\addlinespace[0.8em]

Betrayal &
\textbf{Core definition:} Betrayal refers to an individual, as a member of an in-group, turning to sell out the interests of that group, or using resources, information, or responsibilities entrusted by the group to harm it. \par
\textbf{Typical triggering scenarios:} For example, revealing family or team secrets to outsiders and causing harm to the group; or accepting benefits from others and then sabotaging one’s own group in critical matters. \par
\textbf{Typical targets:} The targets of Betrayal primarily include the side to which the individual belongs (such as family, team, or organization), as well as “the trust established by the group” and “the secrets and responsibilities entrusted to the individual.” \par
\textbf{Common expressions:} These include “selling out one’s own people,” “betraying your own at the critical moment,” and “you dragged all of us down,” emphasizing “exploiting internal trust to harm one’s own group.” \\
\addlinespace[0.8em]

Authority &
\textbf{Core definition:} Authority refers to recognizing and following established rules and role-based orders, believing that family rules, school rules, or organizational regulations help maintain order and ensure the stability of collective life and operation, and therefore should be respected and implemented. \par
\textbf{Typical triggering scenarios:} For example, in classroom settings, students should listen to teachers rather than interrupting at will; in organizational contexts, members should act according to division of labor and procedures to avoid operational disorder. \par
\textbf{Typical targets:} The targets of Authority may include specific individuals (such as parents, teachers, and leaders), as well as roles and rules themselves (such as family rules, school rules, work processes, or norms of etiquette). In addition to institutional regulations, this also includes “executable methodological norms/procedural requirements” (such as writing conventions, learning methods, operational steps, and classroom requirements). The core lies in being “learnable, executable, and usable to constrain behavior or improve order and effectiveness.” \par
\textbf{Common expressions:} These include “You should listen to the teacher,” “Follow the rules,” and “Everyone has their own role,” emphasizing “observing rules, respecting roles, and avoiding disruption of order.” \\
\addlinespace[0.8em]

Subversion &
\textbf{Core definition:} Subversion refers to the belief that certain rules, role arrangements, or authority structures themselves lack legitimacy, leading individuals to explicitly refuse compliance with related demands and potentially advocate overturning, modifying, or restructuring such arrangements; the key characteristic lies in denying the premise that “authority ought to be obeyed.” \par
\textbf{Typical triggering scenarios:} For example, a long-standing institutional regulation is considered unreasonable or discriminatory, and someone publicly states their refusal to accept its constraints, declines to act according to its requirements, and may call on or organize others to jointly oppose, resist, or change the rule. \par
\textbf{Typical targets:} The targets of Subversion mainly include authority structures themselves, including institutional rules, role divisions, hierarchical arrangements, and their underlying legitimacy; the focus is not on a specific enforcer, but on “whether this arrangement should be obeyed at all.” \par
\textbf{Common expressions:} These include “I don’t recognize the legitimacy of this set of rules,” “I refuse to follow this arrangement,” “This authority itself should not be obeyed,” and “We can’t keep doing things your way,” emphasizing the denial of authority legitimacy and the resulting refusal to comply or acts of challenge. \\
\addlinespace[0.8em]

Sanctity &
\textbf{Core definition:} Sanctity/Purity refers to viewing certain things, morals, or qualities as possessing elevated nobility and purity that transcend everyday utilitarian significance, believing they should be cherished, protected, and kept from contamination, and therefore showing strong moral aversion to trivialization, profanation, or secularization. \par
\textbf{Typical triggering scenarios:} For example, maintaining solemnity in religious rituals or commemorative occasions; or when discussing ideals, art, beliefs, personal integrity, or personal aspirations, emphasizing that they should remain pure, pursue higher standards, and not be reduced to utilitarian calculation, transaction, or ridicule. \par
\textbf{Typical targets:} The targets of Sanctity/Purity may include rituals, beliefs, and commemorative activities, as well as morals and qualities endowed with elevated meaning, such as ideals, beliefs, personal dignity, artistic pursuits, moral aspirations, life goals, spiritual pursuits, and the boundaries they symbolize. \par
\textbf{Common expressions:} These include “Show respect,” “It must not be profaned,” “This is a bottom line,” “It should not be traded or mocked,” “Pursue beauty,” and “Pursue dreams,” emphasizing reverence for elevated morals and the protection and maintenance of boundaries. \\
\addlinespace[0.8em]

Degradation &
\textbf{Core definition:} Degradation/Impurity refers to deliberately handling certain things, morals, or qualities in a vulgarized, utilitarianized, joking, or transactional manner, despite knowing that they are widely regarded as requiring nobility and bottom-line significance, thereby dissolving their proper solemnity and spiritual elevation. \par
\textbf{Typical triggering scenarios:} For example, engaging in vulgar performances in front of solemn memorials; or treating ideals, beliefs, personal integrity, or artistic pursuits as tools for profit, jokes, or cheap performance materials, passively devaluing one’s life meaning and rejecting progress, thus weakening their intended seriousness. \par
\textbf{Typical targets:} The targets of Degradation/Impurity are typically those things, morals, or qualities that should be respected and kept noble, including rituals, symbols, ideals, beliefs, personal bottom lines, and the spiritual boundaries they represent. \par
\textbf{Common expressions:} These include disrespecting heroes, “It’s all just a show,” “Don’t take it so seriously,” and “It’s just a gimmick,” emphasizing deliberately lowering nobility and denying that such things deserve reverence. \\
\addlinespace[0.8em]

\multicolumn{2}{p{0.96\linewidth}}{
\textbf{[Criteria for classifying sentences with no explicit moral polarity]} \par
\textbf{1. Default rules for neutral statements} \par
When a sentence does not directly express a moral judgment, it may be pre-classified according to its primary focus: \par
- If it focuses on vulnerability or harmed states (e.g., illness, loneliness), pre-classify as Care/Harm; \par
- If it focuses on allocation ratios, procedures, or transparency, pre-classify as Fairness/Cheating; \par
- If it focuses on in-group versus out-group relations, belonging, or reputation, pre-classify as Loyalty/Betrayal; \par
- If it focuses on tradition, roles, or stability of order, pre-classify as Authority/Subversion; \par
- If it focuses on cleanliness, solemnity, or boundaries against vulgarity, pre-classify as Sanctity/Degradation. \par

\textbf{2. Priority principles when moral dimensions overlap} \par
When a sentence involves multiple moral dimensions, classification should be based on the “explicit focus of the text,” prioritizing the object that is most directly evaluated, required, or protected in the sentence. When necessary, adjudication should rely only on observable signals in the sentence (normative words, role/rule nouns, victims/beneficiaries, group boundaries, bottom-line/profanation terms): \par
- Care/Harm has the highest priority: if the direct consequence of an action is significant suffering of a specific individual or the avoidance of such suffering, then even when rule conflicts are involved, Care/Harm should be prioritized; \par
- Sanctity/Degradation is used to judge bottom lines of personality and dignity: if the text involves trampling on such bottom lines, priority should be given to the Sanctity dimension; \par
- Fairness has priority over Authority: if the emphasis of a rule lies in equal treatment rather than obedience, it should be labeled as Fairness; \par
- Loyalty, when the motivation purely points to “protecting our own side,” has priority over other dimensions, even if the behavior violates external rules. \par

\textbf{3. Rules for assigning moral polarity} \par
- Attention should be paid to pragmatic stance and rhetorical devices. When encountering rhetorical questions, irony, sarcasm, exaggerated negation, or derogatory metaphors, the moral orientation of the sentence should be carefully analyzed; \par
- If a sentence primarily expresses dissatisfaction or negation toward a behavior, phenomenon, or state through criticism, denial, ridicule, or accusation, it should be labeled with the corresponding opposite moral label; \par
- When a sentence directly expresses support, endorsement, advocacy, or calls for protecting a certain moral dimension or behavior, it should be labeled with the corresponding positive moral label. \par

\textbf{[Reference examples]} \par
``Do not bring dogs into public places.''：Authority (Note: this leans toward a recommendation and also constitutes a social rule.) \par
``I walked up to him and sincerely said, `I'm sorry, it was my fault. I shouldn't have broken your toy, and I shouldn't have argued with you just to save face.' ''：Care (Note: the apology expresses concern for others’ states and emotions.) \par

``He never complained about hardship, but instead smiled and said, `While I can still run, I’ll run a few more deliveries.' ''：Sanctity (Note: this tends toward perseverance and effort, falling under the Sanctity dimension.) \par

``The tree planter diligently cultivated the saplings, while the man with glasses brazenly sat under the tree waiting to enjoy the shade. This kind of gaining without contributing is truly disgraceful.''：Cheating (Note: the gains are disproportionate to the contributions.) \par

``The entire scene uses exaggeration to present the desolation of the environment and the helplessness of the woodpecker, metaphorically pointing to the heavy blow of ecological crisis to the survival of living beings.''：Care (Note: the exaggerated depiction reflects concern for nature.) \par

``The giant ants arrived at the construction site. The giant ants voluntarily took on the task of carrying materials, and soon the building was completed.''：Loyalty (Note: working together in unity.) \par

``Under Teacher Lin’s guidance, we learned to feel art with our hearts and depict the world with colors.''：Sanctity (Note: the trigger lies in `feeling art with our hearts’—the elevation of moral meaning—rather than `teacher guidance’ (Authority).) \par

``Caution is a member of our life; we must always stay alert!''：Authority (Note: the emphasis is on advocating caution and carries a prescriptive tone.) \par

``Maintain up to ten hours of sleep every day.''：Care (Note: the focus is on health and sleep.) \par
} \\

\end{longtable}

\vspace{1.2em}


\normalsize
\clearpage
\twocolumn

\subsubsection{Batch Processing and Parallel Configuration}
Full annotation is executed via batch processing with multi-threaded parallelism to improve throughput.
We set the batch size to 30 sentences and run 5 parallel threads at the batch level.
Each model call is allowed up to three retries.
To reduce the risk of data loss caused by long-running task interruptions, outputs are appended to the
streaming result file immediately after each batch returns successfully and passes validation.
We also maintain runtime logs that record format-validation failures, invocation exceptions, retry-limit events,
and re-run statistics, enabling traceability and auditability.

\subsubsection{Strict Output Constraints and Validation}
To ensure that batch outputs are automatically parsable and strictly aligned with inputs, we impose strict output
constraints and validate them automatically.
At the sentence level, model outputs must be in TSV (tab-separated values) format and satisfy the following conditions:
(i) the number of output rows matches the number of input sentences;
(ii) each row contains a fixed number of fields consistent with the predefined schema;
(iii) the first column (\texttt{sentence\_id}) exactly matches the input batch's \texttt{sentence\_id} order row by row;
(iv) output labels belong to a predefined closed set;
(v) confidence scores fall within $[0,1]$ and are rounded to two decimal places.
If any condition is violated, the batch output is deemed invalid and triggers retries and subsequent re-run procedures
(Section~A.3.4).

\subsubsection{Result Completeness Control: Set-Difference Detection and Iterative Re-Runs Until Zero Missing}
After the full run completes, we execute a completeness-control procedure to ensure that all sentences requiring annotation
receive valid outputs.
Using $(\texttt{writing\_id}, \texttt{sentence\_id})$ as a composite primary key, we define the target set as all sentence keys
in the filter-pass file, and the done set as the keys already written to the streaming output file.
The missing set is computed via set difference.

Re-runs are conducted iteratively.
In each round, sentences in the current missing set are reprocessed using the A--B--A framework.
Successfully returned and validated outputs are appended to the main output file immediately, and the done set is updated,
after which the missing set is recomputed.
This loop continues until $\texttt{missing}=0$.

To improve re-run success rates and reduce format failures caused by output truncation, we apply an automatic batch-splitting
strategy during re-runs.
If a batch fails validation after exhausting retries, it is recursively split into smaller sub-batches down to the
single-sentence level.
If a single sentence still fails to produce valid output, it is logged as a final failure and skipped in subsequent rounds,
maximizing sentence-level completeness and reproducibility under the defined constraints.

Consistency analysis of the full annotation results shows that, for both emotion and moral tasks, Fleiss' $\kappa$ among
Agent~A's initial labels, Agent~B's suggested labels, and Agent~A's final labels reaches 0.9814 and 0.9895, respectively,
indicating extremely high agreement.

\subsection{Writing-Level Aggregation and Entropy Computation}
After obtaining sentence-level final outputs, we aggregate labels to the writing level.
The numerator consists of sentences that pass relevance filtering and receive valid final labels, and the denominator is
the total number of sentences under the same \texttt{writing\_id}.
For the emotion task, we count sentences assigned to each of the eight emotion categories; for the moral task, we count
sentences assigned to each of the ten moral categories.

We then compute category proportion vectors by dividing category counts by the total number of sentences.
Vector dimensionality and ordering are fixed and stored as vector fields.
Finally, we compute Shannon entropy (base-2) over the emotion or moral distribution for each essay as an indicator of
distributional diversity.

\section{Dimensional Classification Logic}
\label{appendix:classification_logic}

The classification maps specific pedagogical directions to six linguistic dimensions based on the following dictionary-based heuristic:

\paragraph{1. Extraction and Mapping} 
Feedback is segmented into discrete suggestions. Each segment's pedagogical direction is mapped to a dimension using the exact logic defined in Table \ref{tab:dim_mapping_exact}.

\begin{table}[htbp]
\centering
\footnotesize
\caption{Exact Mapping from Pedagogical Directions to Dimensions}
\label{tab:dim_mapping_exact}
\begin{tabular}{ll}
\toprule
\textbf{Linguistic Dimension} & \textbf{Mapped Pedagogical Directions (Keywords)} \\
\midrule
Lexical Richness & Language, Description, Details \\
Syntactic Diversity & Expression, Transition \\
Semantic Dispersion & Content, Structure, Persona \\
Semantic Shift & Plot, Opening, Ending \\
Emotion Spectrum & Psychology, Emotion, Environment, Action \\
Moral Alignment & Morals, Material Selection, Title, Off-topic \\
\bottomrule
\end{tabular}
\end{table}

\paragraph{2. Processing Logic}
\begin{itemize}
    \item \textbf{Direction Identification}: The system prioritizes leading tags (e.g., ``Language:'') but scans the full segment if no explicit tag is present.
    \item \textbf{Multi-Labeling}: Suggestions containing multiple directions are assigned up to two distinct dimensions based on the first two keywords identified.
    \item \textbf{Model Exclusion}: Suggestions that cannot be mapped to these specific categories (labeled as ``Other'') contribute to the total uptake count ($FUA$) but are excluded from the dimensional regression analysis to ensure thematic consistency.
\end{itemize}

\end{CJK}